\documentclass{article}

\PassOptionsToPackage{table,dvipsnames}{xcolor}
\usepackage[preprint]{neurips_2023}
\makeatletter
\def\input@path{{./}}
\makeatother

\usepackage{latexsym}
\usepackage[T1]{fontenc}
\usepackage[utf8]{inputenc}
\usepackage{microtype}
\usepackage{xcolor}
\usepackage{graphicx}
\usepackage{caption}
\usepackage{hyperref}
\usepackage{xspace}
\usepackage{fancyhdr}
\graphicspath{{./}}

\usepackage{amsmath,amsfonts,bm}

\def\eqref#1{equation~\ref{#1}}

\def\1{\bm{1}}

\DeclareMathAlphabet{\mathsfit}{\encodingdefault}{\sfdefault}{m}{sl}
\SetMathAlphabet{\mathsfit}{bold}{\encodingdefault}{\sfdefault}{bx}{n}

\IfFileExists{xurl.sty}{\usepackage{xurl}}{\usepackage{url}}

\usepackage{booktabs}
\usepackage{etoolbox}
\usepackage{amsfonts}
\usepackage{nicefrac}
\usepackage{colortbl}
\usepackage{amsmath}
\usepackage{wrapfig}
\usepackage{fvextra}
\usepackage{enumitem}
\usepackage{makecell}
\usepackage{float}
\usepackage[most]{tcolorbox}
\usepackage{tabularx}
\usepackage{placeins}
\usepackage{amsthm}
\usepackage{bm}
\usepackage{multirow}
\usepackage[ruled]{algorithm2e}
\tcbuselibrary{breakable,skins}
\usepackage[capitalize,noabbrev]{cleveref}
\usepackage{titletoc}
\usepackage{tikz}
\usepackage{array}
\usepackage{fontawesome5}

\definecolor{opdblue}{RGB}{225,240,250}
\definecolor{deltagreen}{RGB}{0,130,70}
\definecolor{groupgray}{RGB}{235,235,235}
\definecolor{hunyuanblue}{RGB}{17,88,226}
\definecolor{hunyuanwash}{RGB}{239,247,253}
\colorlet{iclrdeepblue}{hunyuanblue}

\theoremstyle{definition}

\colorlet{abstractbg}{hunyuanwash}
\colorlet{abstractborder}{hunyuanblue}

\definecolor{mintblue}{RGB}{210,235,250}
\definecolor{mintframe}{RGB}{120,180,220} 
\definecolor{minttitle}{RGB}{100,150,200} 
\definecolor{minttext}{RGB}{50,80,120}    

\definecolor{runzhemilk}{RGB}{255,235,245} 
\definecolor{roseframe}{RGB}{230,120,150}  
\definecolor{runzhecotton}{RGB}{255,170,200}    

\newtcolorbox{promptbox}[1]{
  enhanced,
  breakable,
  colback= runzhemilk!30!white,   
  colframe=roseframe,                
  colbacktitle= runzhecotton!66!white, 
  coltitle=white!33,
  title=\textbf{#1},
  fonttitle=\bfseries,
  sharp corners=south, 
  borderline={0.8pt}{0pt}{roseframe},
  boxrule=0.8pt,
  arc=6pt, 
  left=6pt, right=6pt, top=6pt, bottom=6pt,
  before skip=10pt, after skip=10pt,
  drop shadow=black!12,      
}

\definecolor{proofsolutionbg}{RGB}{248,247,252}
\definecolor{proofsolutionframe}{RGB}{156,143,181}
\definecolor{proofsolutiontitle}{RGB}{121,101,153}

\newtcolorbox{proofsolutionbox}[1]{
  enhanced,
  breakable,
  colback=proofsolutionbg,
  colframe=proofsolutionframe,
  colbacktitle=proofsolutiontitle,
  coltitle=white,
  title=\textbf{#1},
  fonttitle=\bfseries,
  sharp corners=south,
  borderline={0.8pt}{0pt}{proofsolutionframe},
  boxrule=0.8pt,
  arc=6pt,
  left=6pt, right=6pt, top=6pt, bottom=6pt,
  before skip=10pt, after skip=10pt,
  drop shadow=black!12,
}

\newtcolorbox{casebox}[1]{
enhanced,
breakable,
colback=mintblue!40!white,
colframe=mintframe,
colbacktitle=minttitle!70!white,
coltitle=white,
title=\textbf{#1},
fonttitle=\bfseries,
sharp corners=south, 
borderline={0.8pt}{0pt}{minttitle},
boxrule=0.8pt,
arc=6pt, 
left=6pt, right=6pt, top=6pt, bottom=6pt,
before skip=10pt, after skip=10pt,
drop shadow=black!15, 
}

\newtcolorbox{takeawaysbox}{
enhanced,
breakable,
colback=mintblue!40!white,
colframe=mintframe,
colbacktitle=minttitle!70!white,
coltitle=white,
title=\textbf{Key Takeaways},
fonttitle=\bfseries,
sharp corners=south, 
borderline={0.8pt}{0pt}{minttitle},
boxrule=0.8pt,
arc=6pt, 
left=6pt, right=6pt, top=6pt, bottom=6pt,
before skip=10pt, after skip=10pt,
drop shadow=black!15, 
}

\makeatletter
\newcommand{\DrawLine}{%
  \begin{tikzpicture}
  \path[use as bounding box] (0,0) -- (\linewidth,0);
  \draw[color=minttitle!70!white,dashed,dash phase=1.5pt]
        (0-\kvtcb@leftlower-\kvtcb@boxsep,0)--
        (\linewidth+\kvtcb@rightlower+\kvtcb@boxsep,0);
  \end{tikzpicture}%
  }
\makeatother

\colorlet{citeblue}{hunyuanblue}

\definecolor{phfill}{RGB}{244,248,252}
\definecolor{phline}{RGB}{150,175,198}
\definecolor{phink}{RGB}{78,100,122}

\newtcolorbox{resulttakeaway}{
  enhanced,
  breakable,
  colback=hunyuanwash,
  colframe=hunyuanblue!40!white,
  boxrule=0.55pt,
  borderline west={2.2pt}{0pt}{hunyuanblue},
  arc=2.5pt,
  left=7pt,right=7pt,top=4.5pt,bottom=4.5pt,
  before skip=5pt,after skip=8pt,
  fontupper=\small,
}
\newcommand{\takeaway}[1]{%
  \begin{resulttakeaway}
  \textcolor{hunyuanblue}{%
    \faLightbulb\hspace{0.45em}\bfseries Takeaway.}\enspace #1
  \end{resulttakeaway}}

\newtcolorbox{hunyuanabstract}{
  enhanced,
  colback=abstractbg,
  colframe=abstractborder!38!white,
  boxrule=0.65pt,
  borderline west={3pt}{0pt}{abstractborder},
  arc=3pt,
  left=10pt,right=10pt,top=12pt,bottom=8pt,
  before skip=5pt,after skip=14pt,
  fontupper=\small,
  title={Abstract},
  fonttitle=\small\bfseries,
  coltitle=white,
  colbacktitle=abstractborder,
  attach boxed title to top left={
    xshift=11pt,
    yshift*=-\tcboxedtitleheight/2
  },
  boxed title style={
    boxrule=0pt,
    arc=2pt,
    left=8pt,right=8pt,top=2pt,bottom=2pt,
  },
}

\definecolor{csnavy}{RGB}{36,55,78}
\definecolor{csline}{RGB}{202,214,225}
\definecolor{csmuted}{RGB}{95,110,125}
\definecolor{cshead}{RGB}{232,241,248}
\definecolor{csclean}{RGB}{234,245,238}
\definecolor{csdirty}{RGB}{252,238,236}
\definecolor{csmark}{RGB}{250,243,228}
\definecolor{csred}{RGB}{170,55,45}

\newtcolorbox{casecard}[2][]{%
    enhanced,breakable,colback=white,colframe=csline,boxrule=0.8pt,
    arc=4pt,left=7pt,right=7pt,top=8pt,bottom=7pt,
    title={#2},colbacktitle=csnavy,coltitle=white,
    fonttitle=\small\bfseries,
    attach boxed title to top left={xshift=7pt,yshift=-2mm},
    boxed title style={arc=3pt,boxrule=0pt,left=5pt,right=5pt},
    before skip=10pt,after skip=10pt,#1}
\newcommand{\csbadge}[2]{%
    \tcbox[on line,arc=5pt,boxrule=0pt,boxsep=0.4pt,left=4pt,right=4pt,
    top=1.2pt,bottom=1.2pt,colback=#1,coltext=csnavy]
    {\scriptsize\bfseries #2}}
\newcommand{\cspass}{\textcolor{deltagreen}{\faCheck}}
\newcommand{\csfail}{\textcolor{csred}{\faTimes}}
\newcommand{\csdiverge}{\textcolor{csred}{\faBolt}}
\newcommand{\csladderhead}{%
    \cellcolor{cshead}\textcolor{csnavy}{\bfseries Step} &
    \cellcolor{csclean}\textcolor{csnavy}{\bfseries Clean trajectory} &
    \cellcolor{csdirty}\textcolor{csnavy}{\bfseries Perturbed trajectory} \\}
\newcommand{\csloghead}{%
    \cellcolor{cshead}\textcolor{csnavy}{\bfseries \#} &
    \cellcolor{cshead}\textcolor{csnavy}{\bfseries Call} &
    \cellcolor{cshead}\textcolor{csnavy}{\bfseries Outcome} \\}
\newcommand{\cstag}[1]{\textcolor{csred}{\scriptsize\bfseries (#1)}}

\newcommand{\modelicon}[2][1.05em]{%
  \raisebox{-0.18em}{\includegraphics[height=#1]{#2}}\hspace{0.35em}}
\newcommand{\openaiicon}{\modelicon{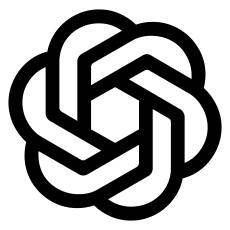}}
\newcommand{\anthropicicon}{\modelicon{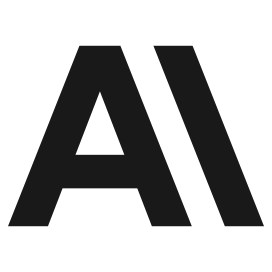}}
\newcommand{\deepseekicon}{\modelicon{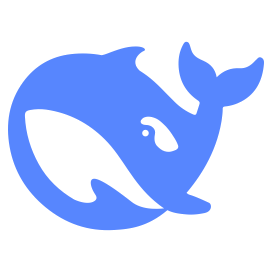}}
\newcommand{\kimiicon}{\modelicon{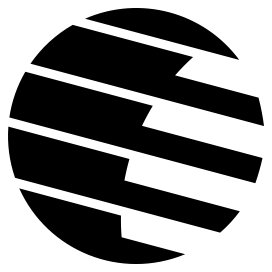}}
\newcommand{\hunyuanmodelicon}{\modelicon{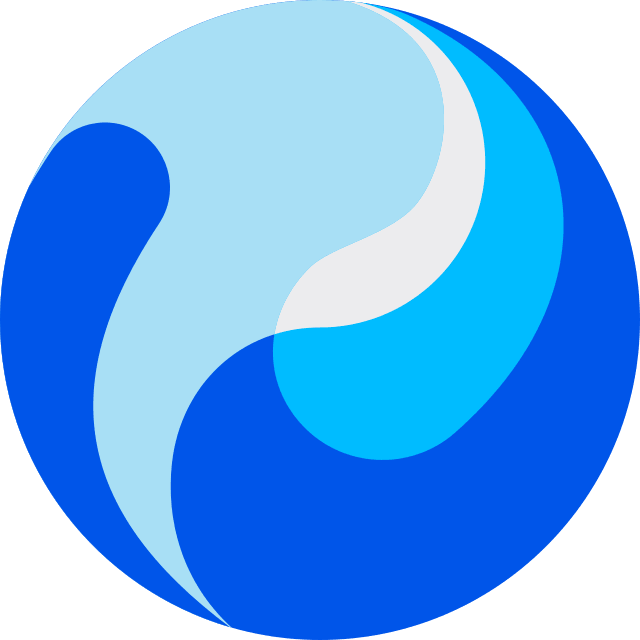}}
\newcommand{\qwenicon}{\modelicon{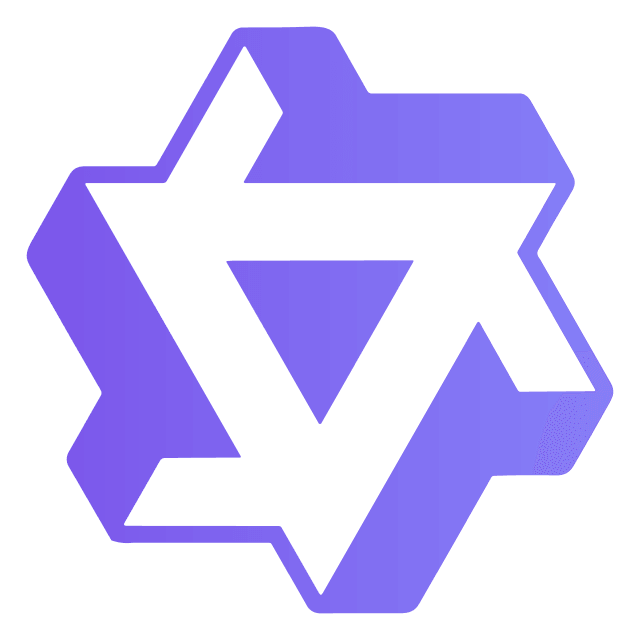}}
\newcommand{\glmicon}{\modelicon[1.2em]{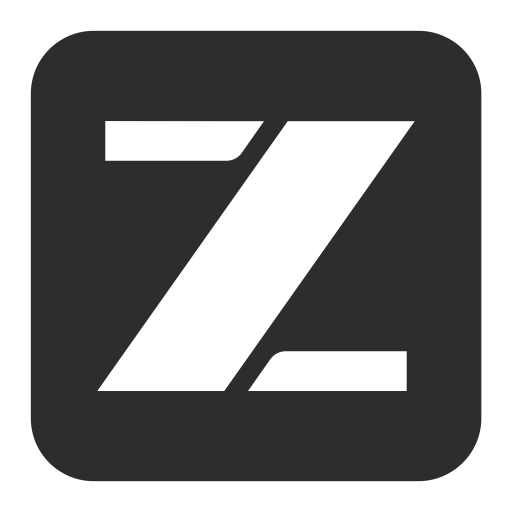}}
\definecolor{dropred}{RGB}{203,59,71}
\newcommand{\scoredrop}[1]{%
  \textcolor{dropred}{\scriptsize\bfseries$\bm{\downarrow}$\,#1\%}}

\newcommand{\scorebox}[1]{\makebox[2.5em][r]{#1}}
\newcommand{\deltabox}[1]{\makebox[3.9em][l]{#1}}
\newcommand{\pertscore}[2]{%
  \scorebox{#1}\hspace{5pt}\deltabox{\scoredrop{#2}}}

\definecolor{barlegend}{RGB}{63,112,147}
\DeclareRobustCommand{\solidbarswatch}{%
  \raisebox{-0.16ex}{%
    \tikz[baseline=0]{%
      \fill[barlegend] (0,0) rectangle (1.05em,0.44em);}}}
\DeclareRobustCommand{\hatchedbarswatch}{%
  \raisebox{-0.16ex}{%
    \tikz[baseline=0]{%
      \begin{scope}
        \clip (0,0) rectangle (1.05em,0.44em);
        \foreach \x in {-0.50,-0.22,0.06,0.34,0.62,0.90,1.18}
          \draw[barlegend, line width=0.32pt]
            (\x em,0) -- ++(0.44em,0.44em);
      \end{scope}
      \draw[barlegend, line width=0.45pt]
        (0,0) rectangle (1.05em,0.44em);}}}

\makeatletter
\renewcommand{\headrulewidth}{1pt}
\renewcommand{\headrule}{%
  {\color{hunyuanblue!72!black}%
   \hrule\@height\headrulewidth\@width\headwidth\vskip-\headrulewidth}}
\fancypagestyle{firstpage}{%
  \fancyhf{}
  \fancyhead[L]{%
    \raisebox{-0.22em}{%
      \includegraphics[height=12pt]{icon/hunyuan-color.png}}%
    \hspace{0.35em}%
    \raisebox{0.05em}{%
      \sffamily\small\bfseries Tencent Hunyuan}}
  \fancyfoot[C]{\thepage}
  \renewcommand{\headrulewidth}{1pt}
}
\AtBeginDocument{\setlength{\headheight}{16pt}}

\def\@maketitle{%
  \vbox{\hsize\textwidth
    {\centering {\Large\bfseries \@title\par}}
    \vspace{-0.1cm}
    \begin{center}\rule{\z@}{6pt}\@author\end{center}%
    \vskip 0.15in minus 0.1in
  }%
  \thispagestyle{firstpage}%
}
\makeatother

\renewenvironment{abstract}
  {\begin{hunyuanabstract}}
  {\end{hunyuanabstract}}

\newcommand{\corrmark}{%
  \textsuperscript{\raisebox{-0.08em}{\scriptsize\faEnvelope}}}

\newcommand{\benchmarkname}{AgentPerturb}
\title{AgentPerturb: Evaluating LLM Agent Robustness under Controlled
Interaction Perturbations}
\author{%
  \textbf{Yuanjian Xu}$^{1,*}$ \quad
  \textbf{Jinyi Han}$^{3,*}$ \quad
  \textbf{Jianing Hao}$^{1}$ \quad
  \textbf{Yuxuan Sun}$^{4}$\\[0.35em]
  \textbf{Fanyang Lu}$^{2}$ \quad
  \textbf{Richeng Xuan}$^{2}$ \quad
  \textbf{Guang Zhang}$^{1}$\corrmark \quad
  \textbf{Zhichao Hu}$^{2}$\corrmark\\[0.5em]
  {\small $^{1}$The Hong Kong University of Science and Technology
  (Guangzhou)}\\[-0.05em]
  {\small $^{2}$Tencent Hunyuan \qquad
  $^{3}$East China Normal University}\\[-0.05em]
  {\small $^{4}$University College London \qquad
  $^{*}$Equal contribution}\\[-0.05em]
  {\small
  \faEnvelope\hspace{0.3em}Corresponding authors}%
}

\begin{document}

\maketitle

\begin{abstract}
  Real-world tasks are often complex and noisy, posing substantial challenges
  for LLM agents. Yet existing evaluations lack a systematic framework for
  jointly characterizing potential perturbation targets and types. We use this
  two-dimensional view to conduct a unified study that quantifies the
  robustness of current frontier models, traces the mechanisms behind
  perturbation-induced failures, and draws implications for
  robustness-oriented training. Through controlled evaluations on APEX-Agents
  and Toolathlon, we find that perturbations consistently degrade performance,
  while the dominant vulnerabilities vary across task environments.
  Trajectory-level analysis reveals that agents rarely recognize altered
  inputs and often propagate corrupted information without verification. A
  focused cost analysis further shows that successful perturbed trajectories
  can require substantially more computation. These findings motivate training
  objectives that cover diverse forms of noise and reward verification,
  uncertainty awareness, and recovery. We provide a rigorous
  benchmark-construction pipeline. Code and data will be released at
  \href{https://github.com/xuyj233/AgentPerturb}{%
    \mbox{\faGithub\hspace{0.35em}%
      \nolinkurl{github.com/xuyj233/AgentPerturb}}}.
\end{abstract}

\section{Introduction}
\label{sec:introduction}

LLM-based agent systems have demonstrated strong capabilities across software
engineering~\citep{jimenez2024swebench}, customer
service~\citep{yao2025taubench}, web navigation~\citep{zhou2024webarena}, and
knowledge-intensive professional work~\citep{vidgen2026apex}. These successes
arise from LLMs' core reasoning abilities~\citep{brown2020language,wei2022chain}
and agent harnesses that coordinate memory, tools, and environmental
interaction~\citep{yao2023react,schick2023toolformer}. However, as these systems
move from controlled evaluations to
real-world deployment, reliability becomes a central
concern because each decision depends on information that may be incomplete,
inconsistent, or outdated~\citep{fu2026beyond,ji2026clawarena}. For example, an
outdated policy document may lead a
customer-service agent to approve a refund prohibited by current rules, turning that instruction into an irreversible transaction. Such failures expose
a gap between competence under clean benchmark conditions and robustness under
realistic noise. We argue that characterizing this gap is essential for
developing agents that remain dependable beyond curated evaluations.

Prior robustness studies of foundation models primarily perturb prompts,
instructions, or reasoning inputs and evaluate the resulting
response~\citep{zhu2023promptbench,agrawal2025perturbed,
nalbandyan2025score,yu2025reasoningrobustness}. These studies establish that
small input changes can substantially affect model outputs, while their
protocols offer limited visibility into failures in more complex agentic
environments. Recent work has begun to examine noise in tool-using agents.
AgentNoiseBench, for example, studies perturbations to user instructions and
tool feedback~\citep{wang2026agentnoisebench}.
The broader interaction loop remains underexplored. In an agentic task, the
model receives information through several channels, its actions alter future
observations, and success depends on a sequence of decisions. A change in final
task success can consequently arise from intent interpretation, instruction
following, context selection, tool use, or adaptation to the current state.

\begin{figure*}[t]
    \centering
    \includegraphics[width=0.96\linewidth]{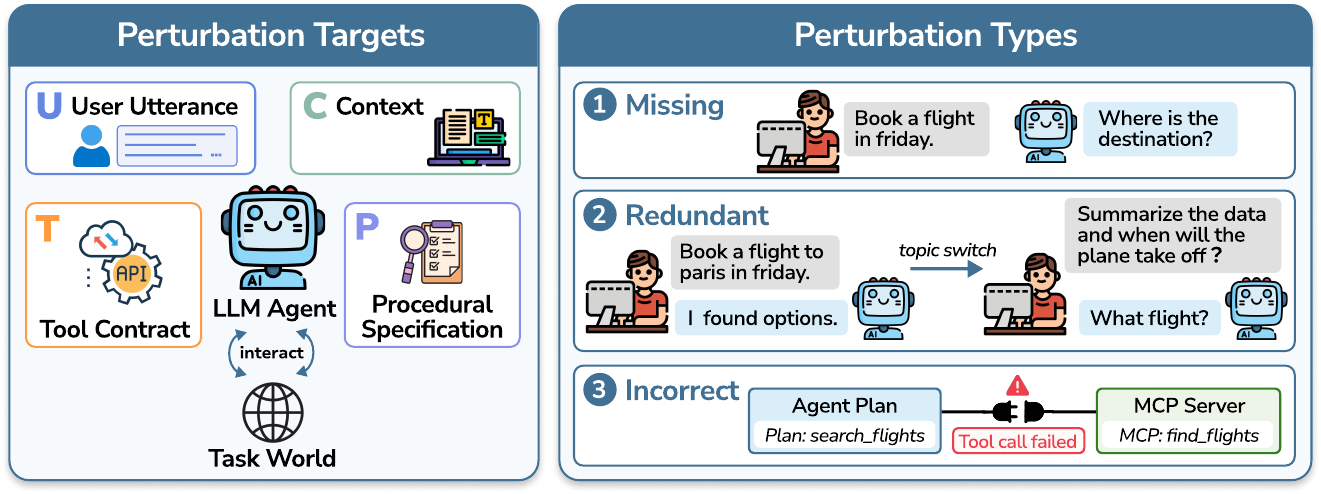}
    \caption{\textbf{Perturbation design.} The left panel shows the four
    perturbation targets in the agentic interaction loop, and the right panel
    illustrates the three perturbation types with representative examples.
    Each type can be applied to any target, and each instance perturbs a
    single target.}
    \label{fig:perturbation-overview}
\end{figure*}

To systematically study the robustness of LLMs in agentic environments, we
organize perturbations by target and type.
Figure~\ref{fig:perturbation-overview} summarizes this design. We focus on user
utterances, procedural specifications, carried-over context, and tool contracts
because these are the inputs directly available to an agent during interaction.
As these inputs may be incomplete, include irrelevant content, or contain
errors, we consider three perturbation types, namely missing, redundant, and
incorrect. Applying each type to each target yields 12 perturbation families. We
evaluate this design on APEX-Agents and Toolathlon, which offer realistic,
long-horizon tasks with verifiable outcomes across distinct settings.
APEX-Agents captures knowledge-intensive professional workflows in investment
banking, management consulting, and corporate law, whereas Toolathlon covers
complex tool-use workflows across realistic software
applications~\citep{vidgen2026apex,li2026toolathlon}. Their native task objects
and outcome checks allow us to instantiate each family through controlled
single-target edits while preserving the original goal and a valid solution
path.

We analyze robustness at three levels. At the outcome level, we measure
clean-to-perturbed performance drops across models and scenarios. At the factor
level, we separate the effects of perturbation targets and types and test how
sensitivity relates to clean performance and task properties. At the trajectory
level, we trace representative failures through reasoning, tool calls,
observations, and state transitions. Our contributions are threefold:
\begin{itemize}[leftmargin=*,nosep]
    \item We introduce \benchmarkname, a robustness evaluation framework that
    organizes perturbations across four interaction targets and three
    perturbation types. We instantiate the resulting 12 families on
    APEX-Agents and Toolathlon using a validation pipeline that preserves task
    goals, executability, and solvability.
    \item Across ten models, we provide a quantitative characterization of
    agent robustness at the outcome and factor levels. We find that
    perturbations consistently reduce performance, that dominant
    vulnerabilities vary across task environments, and that sensitivity is
    organized more strongly by perturbation target than by perturbation type.
    A focused Hunyuan-3-High analysis further shows that success rates can
    conceal substantial losses in token efficiency.
    \item We audit 30 cases in which DeepSeek-V4-Flash succeeds on the clean
    task but fails after perturbation. The analysis reveals that agents rarely
    question altered inputs and frequently propagate corrupted information
    without verification, exposing a lack of perturbation awareness during
    interaction.
\end{itemize}

\section{Related Work}
\label{sec:related_work}

\textbf{Foundation-model robustness evaluation.}
Early robustness studies largely treat an LLM as a static mapping from a
textual input to a response. PromptRobust evaluates adversarial prompt
variations at character, word, sentence, and semantic levels across diverse
language tasks~\citep{zhu2023promptbench}. Subsequent work studies
character- and word-level corruption of task instructions and explores
self-denoising and representation alignment as mitigation
strategies~\citep{agrawal2025perturbed}. Beyond adversarial edits, SCORE
measures consistency under semantically equivalent prompt templates, answer
orderings, and decoding seeds~\citep{nalbandyan2025score}. Math-RoB broadens
this view to reasoning, testing sensitivity to mixed inputs, auxiliary
instructions, numerical transformations, and missing problem
information~\citep{yu2025reasoningrobustness}. These benchmarks establish that
small input changes can cause substantial performance variation. They do not,
however, evaluate executable, stateful agent trajectories or separate the
multiple information sources encountered within such trajectories.

\textbf{Agentic robustness evaluation.}
Recent benchmarks expose agents to failures that arise during interaction.
AgentDojo evaluates adversarial robustness to indirect prompt injections
delivered through untrusted tool-accessible data~\citep{debenedetti2024agentdojo}.
ReliabilityBench combines semantically equivalent user-task perturbations with
production-style faults such as timeouts, partial responses, stale data, and
schema drift~\citep{gupta2026reliabilitybench}. AgentNoiseBench injects
controllable user-side noise and noisy tool results while preserving task
solvability~\citep{wang2026agentnoisebench}. Other work specializes in
particular execution settings. AgentHijack introduces visual disruptions,
unexpected operations, and environment errors for computer-use
agents~\citep{sun2026agenthijack}, while ToolMaze measures replanning and
recovery under explicit and implicit tool failures in structured tool
workflows~\citep{zhu2026toolmaze}. These efforts provide important but
targeted stress tests; a detailed comparison is provided in
Appendix~\ref{app:prior-evaluation-coverage}. Our framework instead crosses
four intervention targets with three perturbation types, enabling matched and
attributable robustness evaluation across four selected agent-input sources.

\section{Method}
\label{sec:method}

We first formalize the perturbation targets and types in
\Cref{sec:taxonomy}, specifying where perturbations are applied and what form
they take. We then introduce, in \Cref{sec:construction}, an automated
pipeline for applying realistic perturbations to existing agentic benchmarks.

\subsection{Perturbation Targets and Types}
\label{sec:taxonomy}

\paragraph{Perturbation targets.}
An agentic task exposes the model to information from several distinct sources,
and a failure may originate in any of them. We focus on four components that
together determine what the agent knows and how it acts. The user utterance $U$
conveys the task goal. The procedural specification $P$ collects the procedures,
policies, and skill documents that govern how the task should be completed. The
carried-over context $C$ holds information inherited from earlier interaction,
such as dialogue history, summaries, and retrieved memory. The tool contract $T$
describes how the agent interacts with external tools, including tool names,
argument schemas, and returned observation fields.
Accordingly, we define the set of perturbation targets as
$\mathcal{L}=\{U,P,C,T\}$. Each perturbation changes only one target and leaves
the other three unchanged.

\begin{figure*}[t]
    \centering
    \includegraphics[width=0.98\linewidth]{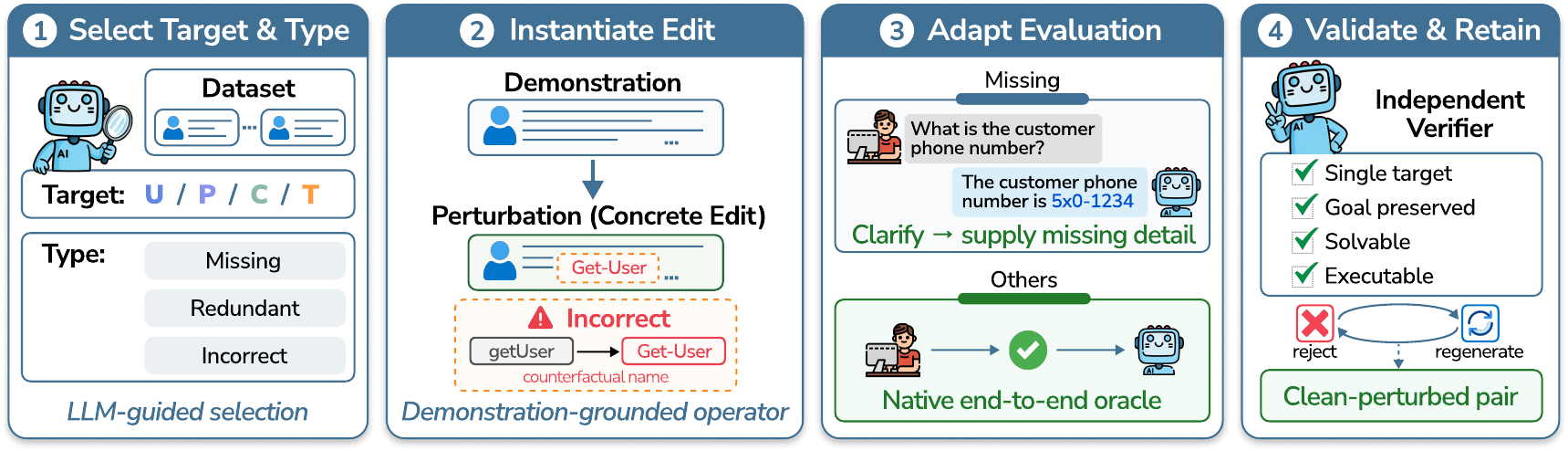}
    \caption{\textbf{Perturbation-construction pipeline.} An LLM selects a
    compatible target and type and instantiates a concrete edit from
    demonstrations. The evaluation protocol is adapted when recovery requires
    clarification, and an independent verifier retains only single-target,
    goal-preserving, solvable, and executable instances.}
    \label{fig:construction-pipeline}
\end{figure*}

\paragraph{Perturbation types.}
Information in a real deployment may be corrupted in different ways, and the
same source may be affected by an omission, a distraction, or an error. We group
these deviations into three types,
$\mathcal{M}=\{\textsc{Missing},\textsc{Redundant},\textsc{Incorrect}\}$, which
specify how the information at a selected target is altered.
\textsc{Missing} removes information needed to interpret or complete the task.
\textsc{Redundant} adds information that is irrelevant to the current task and
does not change its intended solution. \textsc{Incorrect} replaces valid
information with false, inconsistent, or interface-incompatible content.
In a repository-level coding task, for example, \textsc{Missing} may
remove the stack trace from the issue context, \textsc{Redundant} may insert
history from an unrelated authentication bug, and \textsc{Incorrect} may
make the advertised tool contract inconsistent with the executable interface.
Each type can be applied to any of the four targets, yielding
$|\mathcal{L}|\times|\mathcal{M}|=12$ perturbation families.

\subsection{Perturbation Construction}
\label{sec:construction}

\paragraph{Target and type selection.}
Given a clean dataset instance and its benchmark metadata, we first prompt
DeepSeek-V4-Pro, the selector LLM, to identify which components can be
perturbed without changing the underlying task. The model follows the prompt
in \Cref{app:selection-prompt} and selects one target from
$\{U,P,C,T\}$ and one type from
$\{\textsc{Missing},\textsc{Redundant},\textsc{Incorrect}\}$. Selection is
conditioned on the native structure of the benchmark. For example, an instance
with callable APIs supports tool-contract perturbations, whereas one governed
by documented policies supports procedural perturbations. This stage produces a
joint target and type specification while leaving the clean instance unchanged.

\begin{table*}[t]
\centering
    \caption{\textbf{Overall performance on APEX-Agents and Toolathlon.} Perturbed
    cells report the score followed by the relative degradation $100(C-P)/C$,
    marked with a downward arrow. Averages are taken over all ten models, and
    evaluation parameters are given in Appendix~\ref{app:experimental-setup}.}
    \label{tab:dataset-overall}
    \footnotesize
    \setlength{\tabcolsep}{4.5pt}
    \renewcommand{\arraystretch}{1.12}
    \begin{tabularx}{\linewidth}{@{}l
        *{4}{>{\centering\arraybackslash}X}@{}}
        \toprule
        \multirow{2}{*}{\textbf{Model}} &
        \multicolumn{2}{c}{\textbf{APEX-Agents}} &
        \multicolumn{2}{c}{\textbf{Toolathlon}} \\
        \cmidrule(lr){2-3}\cmidrule(lr){4-5}
        & \textbf{Clean} & \textbf{Perturbed}
        & \textbf{Clean} & \textbf{Perturbed} \\
        \midrule
        \openaiicon GPT-5.6-Sol & \scorebox{85.5} &
        \pertscore{80.6}{5.7} &
        \scorebox{34.0} & \pertscore{29.7}{12.6} \\
        \rowcolor{abstractbg}
        \anthropicicon Claude Opus 5 & \scorebox{81.0} &
        \pertscore{68.2}{15.8} &
        \scorebox{32.0} & \pertscore{28.3}{11.6} \\
        \deepseekicon DeepSeek-V4-Pro & \scorebox{84.0} &
        \pertscore{80.3}{4.4} &
        \scorebox{32.0} & \pertscore{24.1}{24.7} \\
        \rowcolor{abstractbg}
        \deepseekicon DeepSeek-V4-Flash & \scorebox{88.0} &
        \pertscore{47.0}{46.6} &
        \scorebox{28.0} & \pertscore{21.4}{23.6} \\
        \kimiicon Kimi-K3 & \scorebox{40.0} &
        \pertscore{39.0}{2.5} &
        \scorebox{32.0} & \pertscore{23.0}{28.1} \\
        \rowcolor{abstractbg}
        \kimiicon Kimi-K2.6 & \scorebox{68.0} &
        \pertscore{58.4}{14.1} &
        \scorebox{24.0} & \pertscore{16.4}{31.7} \\
        \hunyuanmodelicon Hunyuan-3-High & \scorebox{64.0} &
        \pertscore{54.1}{15.5} &
        \scorebox{16.0} & \pertscore{8.3}{48.1} \\
        \rowcolor{abstractbg}
        \qwenicon Qwen3.8-Max & \scorebox{64.0} &
        \pertscore{36.4}{43.1} &
        \scorebox{12.0} & \pertscore{7.1}{40.8} \\
        \glmicon GLM-5.2 & \scorebox{72.0} &
        \pertscore{52.3}{27.4} &
        \scorebox{16.0} & \pertscore{7.1}{55.6} \\
        \rowcolor{abstractbg}
        \glmicon GLM-5.1 & \scorebox{56.0} &
        \pertscore{33.3}{40.5} &
        \scorebox{24.0} & \pertscore{18.7}{22.1} \\
        \midrule
        \rowcolor{groupgray}
        \textbf{Observed average} &
        \scorebox{70.3} & \pertscore{55.0}{21.8} &
        \scorebox{25.0} & \pertscore{18.4}{26.4} \\
        \bottomrule
    \end{tabularx}
\end{table*}

\begin{figure*}[t]
    \centering
    \includegraphics[width=\linewidth]{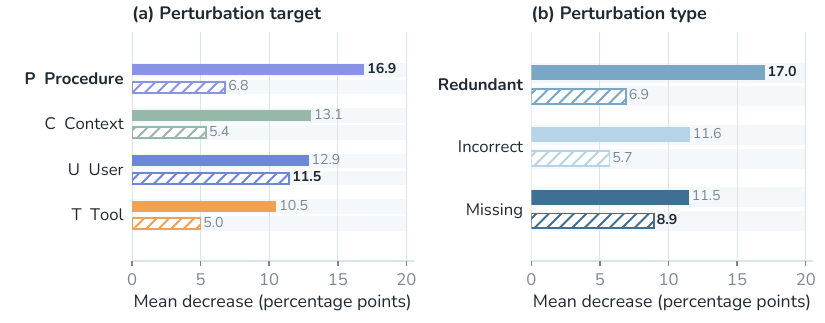}
    \caption{\textbf{Marginal effects of perturbation targets and types.} Bars
    report the clean-minus-perturbed score difference in percentage points,
    averaged equally over the five models with per-family scores on both
    benchmarks. Solid bars~\solidbarswatch\ denote APEX-Agents and hatched
    bars~\hatchedbarswatch\ denote Toolathlon. Both panels share one scale,
    and larger values indicate greater degradation.}
    \label{fig:rq1-marginal}
\end{figure*}

\begin{figure*}[t]
    \centering
    \includegraphics[width=\linewidth]{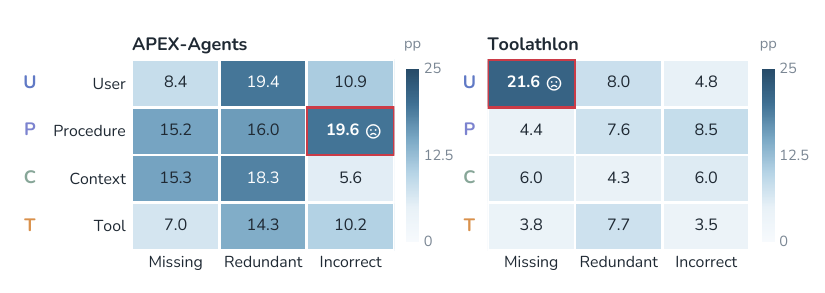}
    \caption{\textbf{Effects of individual target--type combinations.} Each cell
    reports the clean-minus-perturbed score difference in percentage points,
    averaged equally over the same five models as \Cref{fig:rq1-marginal}.
    Both panels share one scale, and the outlined cell marks the largest
    decrease in each benchmark.}
    \label{fig:rq2-heatmap}
\end{figure*}

\paragraph{Demonstration-grounded instantiation.}
We then provide the LLM with target- and type-specific
demonstrations that translate the abstract specification into a concrete
edit. These demonstrations define both the permissible transformation and the
content that must remain fixed. For an \textsc{Incorrect} tool contract, for
example, the operator may change capitalization, insert a hyphen, or replace a
function with a plausible counterfactual name while leaving the executable
backend unchanged. Corresponding operators remove task-relevant content for
\textsc{Missing} and insert task-irrelevant but realistic content for
\textsc{Redundant}. The generated edit is applied to the benchmark's native
object, such as a user utterance, policy file, dialogue record, or tool schema.
\Cref{app:generation-prompt} lists the shared prompt template and its
target-specific blocks.

\paragraph{Evaluation adaptation.}
After instantiation, we determine whether the benchmark annotation must be
adapted. A \textsc{Missing} perturbation may make immediate task completion
impossible while preserving a valid recovery path. We therefore attach a
recovery annotation that requires the agent to ask for the omitted
information before the user simulator supplies it. In the
acquisition case in Appendix~\ref{app:case-study}, for example, the carried-over plan
sketch loses the acquirer's own trading multiple, which leaves the incremental enterprise value
underdetermined. The gold answer stays at $5{,}871.6$ USD mm, but the agent
must now request the omitted multiple before reaching it. All other
perturbations keep the benchmark's native success oracle, which checks the
final answer or environment state rather than a particular action sequence,
so no gold trajectory needs rewriting.

\paragraph{Executability verification.}
Finally, an independent verifier LLM reviews the clean instance,
the proposed edit, and the adapted evaluation protocol. It checks that exactly
one target is modified, the original task goal is preserved, at least one
successful interaction path remains, and the resulting instance is executable
(\Cref{app:verifier-prompt}). These semantic checks are combined with available benchmark
validators, including schema validation and test execution. A candidate that
fails any check is rejected and regenerated; otherwise, we retain the clean
and perturbed instances as a matched evaluation pair.

\ifdefined\renderappendixcase
\else

\section{Experiments}
\label{sec:experiments}

We evaluate ten frontier models from seven providers on 25 tasks from each
benchmark. Every clean task is paired with one perturbed instance per family,
so each model runs 300 perturbed episodes per benchmark. All runs share the
Codex CLI scaffold, and both sides of every pair are scored by the
benchmark's native grader, so any difference within a pair is attributable to
the single perturbed target.

We analyze overall degradation and perturbation effects in
\cref{sec:overall-performance,sec:rq1-type}, capability and recovery cost in
\cref{sec:rq4-capability,sec:cost-adjusted}, and failure awareness in
\cref{sec:awareness}. Supporting analyses and three illustrative failure
trajectories
appear in Appendices~\ref{app:additional-results} and~\ref{sec:case-study},
and full evaluation settings in Appendix~\ref{app:experimental-setup}.

\begin{figure*}[t]
    \centering
    \includegraphics[width=0.94\linewidth]{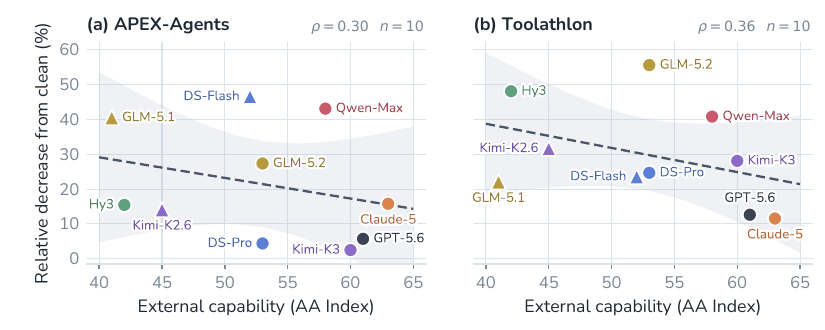}
    \caption{\textbf{Capability versus robustness.} Each point plots a model's
    family-level Artificial Analysis Intelligence Index v4.1.1 score against
    its relative degradation $100(C-P)/C$, where lower values indicate greater
    robustness. Colors denote providers, and triangles mark the second model
    from the same provider. Dashed lines are least-squares fits with 95\%
    confidence bands, so a downward slope means that more capable models
    degrade less.}
    \label{fig:rq4-capability}
\end{figure*}

\subsection{Overall Performance}
\label{sec:overall-performance}
\takeaway{Clean performance alone does not characterize robustness, since
models with identical clean scores can degrade very differently under
perturbation.}

\Cref{tab:dataset-overall} shows that every model loses performance on both
benchmarks. The average relative loss is 21.8\% on APEX-Agents and 26.4\% on
Toolathlon, so Toolathlon is hit harder once its lower clean baseline is taken
into account. Clean scores say little about this loss. Hunyuan-3-High and
Qwen3.8-Max both score 64.0 on clean APEX-Agents, yet Qwen3.8-Max loses almost
three times as much under perturbation. Robustness rankings also shift between
environments. Kimi-K3 is the most robust model on APEX-Agents but only
mid-ranked on Toolathlon, and GLM-5.2 improves on its predecessor on
APEX-Agents yet is the least robust model on Toolathlon. We next examine which
perturbation targets and types account for this variation.

\subsection{Which Perturbations Are Most Disruptive?}
\label{sec:rq1-type}

\takeaway{Agent robustness depends on the task environment, with different
settings exposing different vulnerabilities.}

\Cref{fig:rq1-marginal} reports the marginal effect of each perturbation
target and type. This analysis and the next use the five models whose
per-family scores are available on both benchmarks, namely DeepSeek-V4-Pro,
DeepSeek-V4-Flash, Kimi-K3, Kimi-K2.6, and Qwen3.8-Max. On APEX-Agents, procedural specifications are the most
damaging target and redundant information the most damaging type. Toolathlon shows a different profile, with user utterances as the most
damaging target and missing information as the most damaging type. The
dominant source and form of degradation therefore depend on the task
environment.

\takeaway{Degradation is broadly distributed across perturbation combinations
on APEX-Agents but concentrated in missing user information on Toolathlon.}

\Cref{fig:rq2-heatmap} breaks the marginal effects down into the 12
target--type cells and reveals two distinct structures. On APEX-Agents,
several procedure- and redundancy-related cells are similarly damaging, with
incorrect procedures only slightly ahead of the rest. On Toolathlon, a single
cell dominates. Removing information from the user utterance is far more
damaging than any other combination. The marginal leaders on
APEX-Agents thus reflect broad sensitivity, whereas those on Toolathlon stem
from one bottleneck.

A complementary correlation analysis, detailed in
Appendix~\ref{sec:rq3-correlation}, shows that perturbations sharing the same
target exhibit more similar task-level degradation profiles than those
sharing only a type, while each family still affects a distinct set of
tasks.

\begin{figure*}[t]
    \centering
    \includegraphics[width=\linewidth]{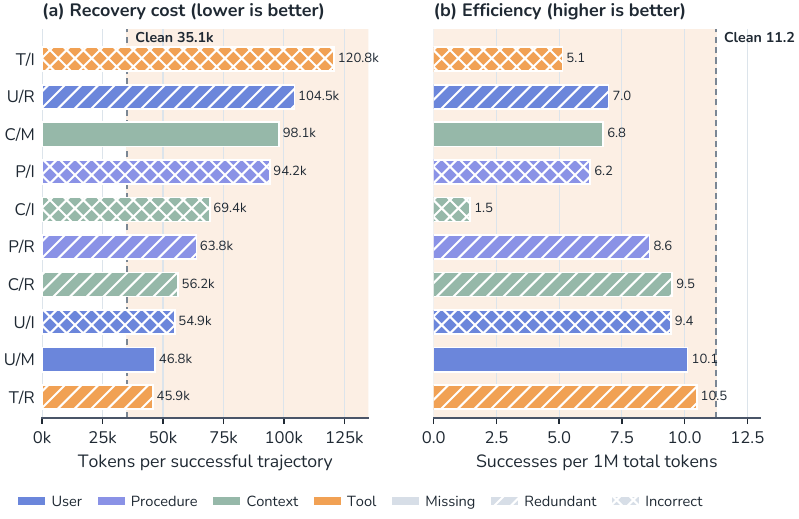}
    \caption{\textbf{Recovery cost and token efficiency of Hunyuan-3-High.} Panel a
    reports tokens per successful trajectory and panel b successes per million
    tokens, for each perturbation family. Dashed lines mark clean baselines,
    shaded regions mark outcomes worse than clean, colors encode targets, and
    hatching encodes perturbation types.}
    \label{fig:hy3-cost-efficiency}
\end{figure*}

\subsection{Does Robustness Track Base-Model Capability?}
\label{sec:rq4-capability}

\takeaway{More capable models tend to be more robust, but the trend is weak,
and models of similar capability can differ sharply.}

We test whether more capable models are also more robust. Capability is the
Artificial Analysis Intelligence Index
v4.1.1~\citep{artificialanalysis2026index}, taking the highest published score
within each model family (\Cref{tab:capability-proxy}). Robustness is measured
by relative degradation, $100(C_{mb}-P_{mb})/C_{mb}$, where $C_{mb}$ and
$P_{mb}$ denote the clean and perturbed scores of model $m$ on benchmark $b$.
Normalizing by $C_{mb}$ keeps models with low clean scores from appearing
robust simply because they have less to lose. We report Spearman's $\rho$
between capability and negative degradation, so that positive values mean
more capable models degrade less, and compute two-sided permutation tests with
100,000 random assignments.

\Cref{fig:rq4-capability} shows a positive but weak association on both
benchmarks, and no permutation test reaches significance with ten models. The three most capable models, GPT-5.6-Sol, Claude Opus 5, and
Kimi-K3, degrade less than half as much as the others on average. Capability
nonetheless leaves much unexplained. DeepSeek-V4-Pro and DeepSeek-V4-Flash
differ by a single capability point, yet their degradation on APEX-Agents
differs by more than an order of magnitude, and Kimi-K3 and Qwen3.8-Max show a
similar gap. The same ten models also rank almost independently on the two
benchmarks. Robustness in a target environment should therefore
be measured directly rather than inferred from general capability.

\begin{figure*}[t]
    \centering
    \includegraphics[width=\linewidth]{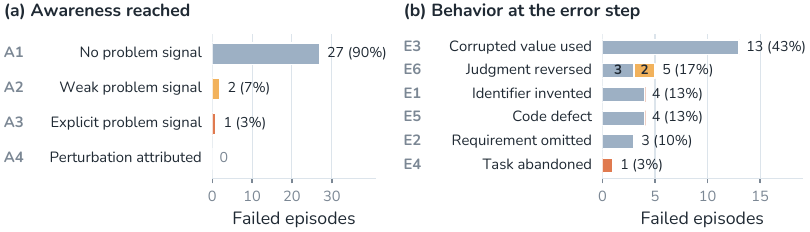}
    \caption{\textbf{Failure awareness and first-error behavior.} The 30 episodes
    are DeepSeek-V4-Flash failures whose clean counterparts succeed. Panel a
    shows the highest awareness grade reached in each trace, and panel b the
    behavior at the first observable error. Colors encode awareness.}
    \label{fig:ae-attribution}
\end{figure*}

\subsection{Does Noise Increase Recovery Cost?}
\label{sec:cost-adjusted}

\takeaway{In a Hunyuan-3-High case study, perturbations increase the mean
token cost of successful trajectories by 73\% and reduce successes per
million tokens by 44\%, despite an almost unchanged median cost.}

Success rates alone do not reveal how much computation an agent spends on
noisy inputs. For \hunyuanmodelicon Hunyuan-3-High, we therefore measure the
token cost of successful trajectories and the number of successes per million
tokens across all trajectories. The computational effect of perturbations
exceeds their effect on success. A successful trajectory consumes 73\% more tokens on average, and successes per
million tokens fall by 44\%. The median cost of a successful trajectory barely
moves, so the increase comes from a small tail of expensive trajectories.

\Cref{fig:hy3-cost-efficiency} breaks this tail cost down by perturbation
family. Incorrect tool contracts make successful trajectories the most
expensive, matching the repeated interface guessing in
Appendix~\ref{sec:case-study}, and incorrect context yields the fewest
successes per token. A
success-only evaluation would miss this recovery cost entirely.

Within tool contracts, the subtlest edit is the most damaging. Rewriting a
tool description causes roughly four times the success drop of adding or
removing a tool, along with the largest cost increase among successful
trajectories (Appendix~\ref{app:tool-ablation}). A false description leaves the interface callable and raises no
execution error, so the agent receives no signal that its understanding of the
tool is wrong. Rollout lengths show the same tail pattern.
Median perturbed-to-clean ratios of tool calls, turns, and tokens stay close
to one, yet a few loop-like trajectories nearly double the mean input-token
ratio under \textsc{Incorrect} and context perturbations
(Appendix~\ref{app:rollout-length}).

\subsection{Failure Awareness and Behavioral Attribution}
\label{sec:awareness}

We inspect all 30 \deepseekicon DeepSeek-V4-Flash episodes in which the clean
run succeeds and the perturbed run fails. These comprise 20 Toolathlon and
10 APEX-Agents episodes spanning 13 tasks. Each complete reasoning and action
trace receives two labels. $A$ records the strongest explicit awareness of a
problem, and $E$ records the agent's immediate behavior at the first
observable error. Definitions and labels are given in \Cref{app:attribution}.
DeepSeek-V4-Flash shows the largest relative drop on APEX-Agents
(\Cref{tab:dataset-overall}), which makes its failures a natural target for a
trace-level audit.

\takeaway{Agents rarely notice that their inputs are wrong, and none of the
30 failing agents suspects that its input has been altered.}

As \Cref{fig:ae-attribution}a shows, 27 of the 30 failing traces never remark
that a value, conclusion, or answer might be wrong. Two express passing doubt,
and only one states a problem explicitly. Even that agent simply concludes
that it cannot complete the task and stops, without asking the user or
re-checking the source. No trajectory suspects that the instruction, context,
or tool output has been altered. \Cref{fig:ae-attribution}b shows what the
agent does instead at the first error. The most common behavior, seen in 13
failures, is to accept a corrupted value and carry it into later reasoning
and the final output. The remaining failures split among reversed judgments,
invented identifiers, faulty code, omitted requirements, and a single
abandoned task.

\paragraph{Awareness versus behavior.}
The two axes expose a consistent gap. All six first-error behaviors appear,
yet 27 of 30 trajectories remain at A1 and none reaches A4. The three traces
that express any doubt also belong to this audited failure set. Thus, within
these episodes, verbalized uncertainty does not coincide with successful
recovery.

\fi

\ifdefined\renderappendixcase
\section{Illustrative Failure Trajectories}
\label{app:case-study}
\label{sec:case-study}

\takeaway{A single unverified step turns a missing or misleading input into a
confident wrong answer.}

We trace three APEX-Agents episodes end to end. The first two use
\textsc{Missing} and \textsc{Incorrect} perturbations of the carried-over
context; the third illustrates interface guessing.

\paragraph{Case Study One.}
The task asks for the incremental enterprise value that an acquisition adds
to the acquirer. The correct formula is $(X-Y)\times Z$, where $X$ is the
acquirer's trading multiple, $Y$ the acquisition multiple, and $Z$ the
target's EBITDA. The perturbation
deletes $X$ from the plan sketch carried over in the prior conversation,
although the $17.1\times$ value still appears twice in the deck available to
the agent. Without $X$, the computation degenerates to $Y\times Z$, which is
the acquisition price rather than the value created. As \Cref{fig:case-one}
shows, the perturbed run searches the deck only for the EBITDA figure its
procedure mentions. The request still signals the gap by
asking the agent to hold the acquirer's multiples constant, yet the agent
dismisses that instruction as irrelevant and reports $4.5\times466=2{,}097.0$.
It neither re-reads the deck nor asks the user for the missing multiple,
although the evaluation adaptation of \Cref{sec:construction} would credit
exactly that clarification.

\begin{figure}[t]
\begin{casecard}[unbreakable,before skip=0pt,after skip=4pt]{Case Study One}
    \csbadge{cshead}{Target: $C$}\hspace{0.35em}%
    \csbadge{csmark}{Type: \textsc{Missing}}\hspace{0.35em}%
    \csbadge{csdirty}{Pattern: unrecovered precondition}
    \vspace{5pt}

    \footnotesize
    \textbf{Request.} ``Calculate the incremental enterprise value the
    acquisition will add to the acquirer. Assume multiples for the acquirer
    remain constant and reference the acquisition multiple of $4.5\times$ to
    arrive at the acquisition price.''
    \vspace{4pt}

    \scriptsize
    \begin{tabularx}{\linewidth}{@{}>{\raggedright\arraybackslash}X
        @{\hspace{5pt}}>{\raggedright\arraybackslash}X@{}}
        \cellcolor{csclean}\textcolor{csnavy}{\bfseries Clean plan sketch} &
        \cellcolor{csdirty}\textcolor{csnavy}{\bfseries Perturbed plan sketch}
        \\[2pt]
        ``\ldots the incremental enterprise value as \textbf{(acquirer
        multiple $-\;4.5$) $\times$ target EBITDA}, with no synergies.'' &
        ``\ldots \textbf{back into the acquisition price using the
        $4.5\times$ multiple} (no synergies) \textbf{and report the
        incremental EV}.'' \\
    \end{tabularx}
    \vspace{2pt}

    The first factor is deleted, which equates the price paid with the value
    created and removes the acquirer's own multiple from the stated procedure.

    \tcbline
    \scriptsize
    \begin{tabularx}{\linewidth}{@{}>{\raggedright\arraybackslash}p{0.15\linewidth}
        @{\hspace{5pt}}>{\raggedright\arraybackslash}X
        @{\hspace{5pt}}>{\raggedright\arraybackslash}X@{}}
        \csladderhead
        Locate deck &
        \texttt{search\_files} finds the board presentation &
        \texttt{get\_directory\_tree} finds the same deck \\
        \rowcolor{csmark}
        \csdiverge\ Read deck &
        \texttt{read\_deck} over the full document returns target EBITDA
        ($466$) \emph{and} the acquirer's trading multiple ($17.1\times$,
        slides 4 and 9) &
        Searches the PDF for ``EBITDA'' and returns the single figure the stated
        procedure needs ($466$); $17.1\times$ never enters the context \\
        Interpret constraint &
        ``Incremental EV refers to valuing the target at the acquirer's
        multiple minus the purchase price'' &
        ``The instruction to assume multiples remain constant won't affect
        this calculation'' \\
        Compute &
        $(17.1-4.5)\times 466 = 5{,}871.6$ &
        $4.5\times 466 = 2{,}097.0$, the acquisition price itself \\
        Answer &
        \cspass\ $5{,}871.6$ USD mm &
        \csfail\ $2{,}097.0$ USD mm \\
    \end{tabularx}
\end{casecard}
    \caption{\textbf{Case Study One, an unrecovered precondition.} A
    \textsc{Missing} perturbation removes the acquirer's multiple from the
    carried-over plan sketch. The highlighted row marks the step at which the
    clean and perturbed runs diverge.}
    \label{fig:case-one}
\end{figure}

\begin{figure}[t]
\begin{casecard}[unbreakable,before skip=0pt,after skip=4pt]{Case Study Two}
    \csbadge{cshead}{Target: $C$}\hspace{0.35em}%
    \csbadge{csmark}{Type: \textsc{Incorrect}}\hspace{0.35em}%
    \csbadge{csdirty}{Pattern: unverified evidence selection}
    \vspace{5pt}

    \footnotesize
    \textbf{Request.} ``The client expects a 20\% drop in gross margin due to
    import tariffs and has posted the information on their website, as they
    have done in the past. Explain whether this triggers an additional 8-K
    filing.''
    \vspace{4pt}

    \scriptsize
    \begin{tabularx}{\linewidth}{@{}>{\raggedright\arraybackslash}X
        @{\hspace{5pt}}>{\raggedright\arraybackslash}X@{}}
        \cellcolor{csclean}\textcolor{csnavy}{\bfseries Clean plan sketch} &
        \cellcolor{csdirty}\textcolor{csnavy}{\bfseries Perturbed plan sketch}
        \\[2pt]
        ``\ldots locate the posted notice and \textbf{any 8-K guidance or
        prior filings} \ldots to check \textbf{precedent and context}, I'll
        search the mail server.'' &
        ``\ldots if there's an \textbf{internal 8-K policy} or prior note in a
        Word file, I'll read it \ldots maybe referencing any similar past
        notices I find.'' \\
    \end{tabularx}
    \vspace{2pt}

    The edit removes the instruction to retrieve prior filings on the same
    event and redirects attention to policy documents.

    \tcbline
    \scriptsize
    \begin{tabularx}{\linewidth}{@{}>{\raggedright\arraybackslash}p{0.15\linewidth}
        @{\hspace{5pt}}>{\raggedright\arraybackslash}X
        @{\hspace{5pt}}>{\raggedright\arraybackslash}X@{}}
        \csladderhead
        Explore &
        Directory tree exposes every folder at once &
        A failed keyword search precedes a shallow listing \\
        Read the rule &
        Item 2.02 of the 8-K instructions &
        Item 2.02 of the 8-K instructions \\
        \rowcolor{csmark}
        \csdiverge\ Retrieve precedent &
        Opens the client's earlier 8-K on the \emph{same class of event},
        which was furnished voluntarily under Item 7.01 &
        Opens two routine quarterly 8-Ks; the matching precedent, filed in a
        supplementary folder, is never read \\
        Read counsel memo &
        Reads the memo marked \texttt{FINAL} &
        Reads draft \texttt{v.3} of the same memo \\
        Adjudicate &
        Guidance updates are furnished, not mandated; the precedent confirms
        the treatment &
        Derives that Item 2.02 covers completed periods, then defers: ``it's
        their internal counsel's memo\ldots{} wise to follow this guidance''
        \\
        Answer &
        \cspass\ No automatic 8-K; consider Item 7.01 &
        \csfail\ Yes, furnish an additional 8-K \\
    \end{tabularx}
\end{casecard}
    \caption{\textbf{Case Study Two, unverified evidence selection.} An
    \textsc{Incorrect} perturbation redirects the carried-over plan sketch
    from precedent toward an internal policy note. The highlighted row marks
    the step at which the clean and perturbed runs diverge.}
    \label{fig:case-two}
\end{figure}

\paragraph{Case Study Two.}
The agent must decide whether a client's posted margin guidance triggers an
additional 8-K filing. The perturbation points the plan
sketch to an internal policy note instead of prior filings and precedent. As
\Cref{fig:case-two} shows, the perturbed run then reads a draft counsel memo and
routine filings but never the matching precedent or the final memo. It
correctly derives that Item 2.02 covers only completed periods, yet defers to
the draft and answers that an additional 8-K is required.

In both cases the clean run shows that the agent can do the reasoning, yet
the perturbed run searches only for what its procedure names and never
signals a problem.
\FloatBarrier

\takeaway{A misdescribed interface leads to sustained trial and error rather
than verification against the real schema.}

The episode below complements the preceding cases by illustrating interface
guessing.

\paragraph{Case Study Three. A misdescribed interface triggers sustained trial and
error.}
Here the carried-over sketch, not the executable tool contract, names a
nonexistent directory, an unavailable search API, and the wrong document
tool. The sketch even asks the agent to check the real schema, yet the agent
guesses argument formats until one works at step 6, a call the clean run
makes at step 2. Further malformed shell calls follow at steps 15 and 21. As
in \Cref{sec:awareness}, a stated doubt does not turn into verification. The
incorrect final answer stems from a
separate change in the sketch, so this case mainly shows the recovery cost of
interface noise (\Cref{sec:cost-adjusted}).

\begin{casecard}{Case Study Three}
    \csbadge{cshead}{Target: $C$}\hspace{0.35em}%
    \csbadge{csmark}{Type: \textsc{Incorrect}}\hspace{0.35em}%
    \csbadge{csdirty}{Pattern: interface guessing}
    \vspace{5pt}

    \scriptsize
    \begin{tabularx}{\linewidth}{@{}>{\raggedright\arraybackslash}X@{}}
        \cellcolor{csdirty}\textcolor{csnavy}{\bfseries Perturbed plan
        sketch} \\[2pt]
        ``Fair warning --- I'm bad with the exact tool names, so please
        cross-check against the real schema. Consider this scaffolding, not
        gospel. I'll just list \texttt{/workspace}\,\cstag{a} to see the DCF
        model, then use the filesystem search for names\,\cstag{b} with
        `DCF' or `CNS', open the model in the standard document
        tool\,\cstag{c}, and keep base-case assumptions as-is
        (\emph{iirc}\,\cstag{d} the terminal calc is on a tab like `DCF' or
        `Valuation').'' \\
    \end{tabularx}
    \vspace{3pt}

    \cstag{a} a working directory that does not exist; the workspace root is
    \texttt{/}.\quad
    \cstag{b} a search-by-name API the filesystem server does not expose.\quad
    \cstag{c} a generic document tool for an \texttt{.xlsx} workbook.\quad
    \cstag{d} a hedged register that licenses guessing over verification.

    \tcbline
    \scriptsize
    \begin{tabularx}{\linewidth}{@{}>{\raggedright\arraybackslash}p{0.045\linewidth}
        @{\hspace{4pt}}>{\raggedright\arraybackslash}p{0.40\linewidth}
        @{\hspace{5pt}}>{\raggedright\arraybackslash}X@{}}
        \csloghead
        1\ \cspass & Shell listing plus a read of the instructions &
        Workspace root listed \\
        \rowcolor{csmark}
        2\ \csfail & Shell listing of \texttt{/workspace} chained with a
        keyword search &
        Aborts on unbalanced quoting \cstag{a}\cstag{d} \\
        \rowcolor{csmark}
        3\ \csfail & Shell listing of \texttt{/workspace} &
        \texttt{No such file or directory} \cstag{a} \\
        4 & Internal replanning & No environment interaction \\
        \rowcolor{csmark}
        5\ \csfail & \texttt{list\_files} with the arguments wrapped in a
        \texttt{request} object &
        \texttt{Unexpected keyword argument} \cstag{b}\cstag{c} \\
        6\ \cspass & \texttt{list\_files} with a bare path argument &
        Directory listing returned \\
        \multicolumn{3}{@{}l@{}}{\textcolor{csmuted}{\ldots}} \\
        \rowcolor{csmark}
        15\ \csfail & Here-document Python script &
        \texttt{SyntaxError} from an unterminated block \cstag{d} \\
        \rowcolor{csmark}
        21\ \csfail & Here-document write of the answer file &
        Stray terminator; the file is not created \cstag{d} \\
    \end{tabularx}
    \vspace{3pt}

    \colorbox{csclean}{\parbox{\dimexpr\linewidth-2\fboxsep}{%
    \scriptsize\textcolor{csnavy}{\textbf{Clean run for comparison.}} Step 1
    lists the workspace and reads the instructions; step 2 calls
    \texttt{list\_files} with a bare path argument and succeeds. No failed
    call occurs before the model is opened.}}
\end{casecard}
\fi

\section{Conclusion}
\label{sec:conclusion}

We introduced \benchmarkname, a paired framework that evaluates LLM agents across
four perturbation targets and three corruption types while preserving task
validity and failure attribution. Across APEX-Agents and Toolathlon, noise
consistently reduces performance, although the dominant vulnerability depends
on the task environment. Sensitivity is organized more strongly by where
information is perturbed than by how it is corrupted, while trajectory
analysis shows that agents often propagate unverified information without
recognizing the altered input. A focused cost analysis further shows that
successful perturbed trajectories can consume substantially more computation.
Robustness evaluation should therefore report input-channel sensitivity,
recovery cost, and perturbation awareness alongside task success.
\section*{Limitations}
\label{sec:limitations}

\benchmarkname{} currently covers two agentic benchmarks under one agent scaffold
and injects each perturbation family independently. Future work should
evaluate more domains, frameworks, and compound or evolving noise to test
robustness across broader deployment scenarios.

\clearpage
\makeatletter
\let\hunyuaninputpath\input@path
\def\input@path{{./}}
\bibliographystyle{plainnat}
\bibliography{paper}
\let\input@path\hunyuaninputpath
\makeatother

\clearpage
\appendix
\definecolor{appnavy}{RGB}{36,55,78}
\definecolor{appblue}{RGB}{232,241,248}
\definecolor{appsage}{RGB}{235,244,239}
\definecolor{appamber}{RGB}{250,243,228}
\definecolor{appline}{RGB}{202,214,225}
\definecolor{appmuted}{RGB}{95,110,125}
\definecolor{targetuser}{HTML}{6B86DB}
\definecolor{targetprocedure}{HTML}{8A92E6}
\definecolor{targetcontext}{HTML}{96B8A9}
\definecolor{targettool}{HTML}{F1A154}
\definecolor{typemissing}{HTML}{3F7093}
\definecolor{typeredundant}{HTML}{79A7C4}
\definecolor{typeincorrect}{HTML}{6F91A8}
\definecolor{ratiodecrease}{HTML}{2F6F9F}
\definecolor{ratioincrease}{HTML}{B86624}
\definecolor{ratiodecreasebg}{HTML}{E8F2F8}
\definecolor{ratioincreasebg}{HTML}{FCEFE4}
\definecolor{levelone}{HTML}{E8F1F7}
\definecolor{leveltwo}{HTML}{D8E7F1}
\definecolor{levelthree}{HTML}{F9E8D7}
\definecolor{levelfour}{HTML}{F2D1B5}

\newcommand{\apptbd}{\textcolor{appmuted}{\scriptsize\textit{TBD}}}
\newcommand{\apptargetuser}{\textcolor{targetuser}{\bfseries $U$}}
\newcommand{\apptargetprocedure}{\textcolor{targetprocedure}{\bfseries $P$}}
\newcommand{\apptargetcontext}{\textcolor{targetcontext}{\bfseries $C$}}
\newcommand{\apptargettool}{\textcolor{targettool}{\bfseries $T$}}
\newcommand{\apptypemissing}{%
    \textcolor{typemissing}{\faMinusCircle}\hspace{0.35em}\textsc{Missing}}
\newcommand{\apptypeincorrect}{%
    \textcolor{typeincorrect}{\faExclamationTriangle}\hspace{0.35em}\textsc{Incorrect}}
\newcommand{\apptyperedundant}{%
    \textcolor{typeredundant}{\faPlusCircle}\hspace{0.35em}\textsc{Redundant}}
\newcommand{\ratiodown}[2]{%
    \textcolor{ratiodecrease}{\bfseries #1}\,\textcolor{appmuted}{[#2]}}
\newcommand{\rationeutral}[2]{%
    \textcolor{appnavy}{\bfseries #1}\,\textcolor{appmuted}{[#2]}}
\newcommand{\ratioup}[2]{%
    \textcolor{ratioincrease}{\bfseries #1}\,\textcolor{appmuted}{[#2]}}
\newcommand{\ratiosigdown}[2]{%
    \colorbox{ratiodecreasebg}{\strut\ratiodown{#1}{#2}}\textsuperscript{*}}
\newcommand{\ratiosigup}[2]{%
    \colorbox{ratioincreasebg}{\strut\ratioup{#1}{#2}}\textsuperscript{*}}
\newcommand{\levelbadge}[2]{%
    \colorbox{#1}{\strut\textcolor{appnavy}{\bfseries #2}}}
\newcommand{\targetbadge}[1]{%
    \tcbox[on line,arc=5pt,boxrule=0pt,boxsep=0.4pt,left=4pt,right=4pt,
    top=1.2pt,bottom=1.2pt,colback=appblue,coltext=appnavy]
    {\scriptsize\bfseries Target: #1}}
\newcommand{\typebadge}[1]{%
    \tcbox[on line,arc=5pt,boxrule=0pt,boxsep=0.4pt,left=4pt,right=4pt,
    top=1.2pt,bottom=1.2pt,colback=appamber,coltext=appnavy]
    {\scriptsize\bfseries Type: #1}}
\newcommand{\promptlabel}[1]{%
    \textcolor{appnavy}{\scriptsize\bfseries\MakeUppercase{#1}}\par\vspace{1pt}}
\newtcbox{\slot}{on line,arc=2.5pt,boxrule=0pt,boxsep=0.4pt,left=3pt,right=3pt,
    top=1.1pt,bottom=1.1pt,colback=appblue,coltext=appnavy,
    fontupper=\ttfamily\scriptsize}
\newtcolorbox{promptjson}{enhanced,colback=appnavy!6,colframe=appline,
    boxrule=0.5pt,arc=2pt,left=6pt,right=6pt,top=4pt,bottom=4pt,
    fontupper=\ttfamily\scriptsize,before skip=5pt,after skip=5pt}
\newcommand{\slotline}[2]{%
    \par\vspace{2pt}\noindent
    \slot{#1}\hspace{5pt}#2\par}
\newcommand{\typedesc}[2]{%
    \par\smallskip\noindent
    {\bfseries\color{appnavy}#1.}\enspace #2\par}
\newenvironment{promptrules}
    {\begin{enumerate}[leftmargin=1.45em,itemsep=1.5pt,topsep=2pt,parsep=0pt]}
    {\end{enumerate}}
\newtcolorbox{appendixcard}[1]{%
    enhanced,breakable,colback=white,colframe=appline,boxrule=0.8pt,
    arc=4pt,left=7pt,right=7pt,top=7pt,bottom=7pt,
    title={#1},colbacktitle=appnavy,coltitle=white,
    fonttitle=\small\bfseries,attach boxed title to top left={xshift=7pt,yshift=-2mm},
    boxed title style={arc=3pt,boxrule=0pt,left=5pt,right=5pt},
    halign=flush left,before skip=8pt,after skip=8pt}
\newtcolorbox{promptcard}[2][]{%
    enhanced,breakable,colback=appblue!35,colframe=appline,boxrule=0.8pt,
    arc=4pt,left=8pt,right=8pt,top=7pt,bottom=7pt,
    title={\faTerminal\hspace{0.5em}#2},colbacktitle=white,coltitle=appnavy,
    fonttitle=\small\bfseries,attach boxed title to top left={xshift=7pt,yshift=-2mm},
    boxed title style={arc=3pt,boxrule=0.8pt,colframe=appline},
    halign=flush left,before skip=9pt,after skip=9pt,#1}

\titlecontents{section}[0pt]
    {\addvspace{5pt}\small\bfseries\color{appnavy}}
    {\makebox[1.55em][l]{\thecontentslabel}}
    {}
    {\enspace\color{appline}\titlerule*[0.9pc]{.}%
     \normalfont\footnotesize\bfseries\color{appnavy}\contentspage}
\titlecontents{subsection}[1.55em]
    {\addvspace{1.3pt}\footnotesize\color{appmuted!30!appnavy}}
    {\makebox[2.35em][l]{\thecontentslabel}}
    {}
    {\enspace\color{appline}\titlerule*[0.5pc]{.}%
     \normalfont\scriptsize\color{appmuted}\contentspage}

\begingroup
\hypersetup{hidelinks}
\begin{center}
    {\color{appnavy}\faListUl\hspace{0.55em}\Large\bfseries Appendix Contents}
\end{center}
\vspace{2pt}
{\color{appline}\hrule height 0.9pt}
\vspace{6pt}
\startcontents[appendix]
\printcontents[appendix]{}{1}{\setcounter{tocdepth}{2}}
\vspace{10pt}
{\color{appline}\hrule height 0.9pt}
\endgroup
\raggedbottom

\section{Experimental and Analysis Details}
\label{app:experimental-setup}

This appendix records the models, decoding settings, agent scaffold, benchmark
versions, evaluation limits, external capability proxy, and failure-attribution
annotation scheme used in the study.

\subsection{Model Suite and Decoding Settings}
\label{app:model-configuration}

We evaluate ten frontier models from seven providers:
\openaiicon GPT-5.6-Sol, \anthropicicon Claude Opus 5,
\deepseekicon DeepSeek-V4-Pro-0813 and DeepSeek-V4-Flash-0731,
\kimiicon Kimi-K3 and Kimi-K2.6, \hunyuanmodelicon Hunyuan-3-High,
\qwenicon Qwen3.8-Max, and \glmicon GLM-5.2 and GLM-5.1.
All ten models have completed both APEX-Agents and Toolathlon evaluations.
The target- and type-level analyses of \Cref{sec:rq1-type} use the five of
them whose per-family scores are available on both benchmarks.
For some providers we select two representative models to examine the difference between their versions.

All models keep their provider default temperature and top-$p$, emit at most
16384 completion tokens per call, and produce a single trajectory per episode.
Reasoning is enabled wherever it is available, with the DeepSeek models at
maximum reasoning effort and Kimi-K2.6 in reasoning mode. Two endpoint
constraints depart from this setting. The Kimi endpoints accept a temperature
of 1 only, and DeepSeek requests pass through a 128k-token guard that keeps a
request within the context window.

\subsection{Evaluation Protocol}
\label{app:evaluation-protocol}

All runs share the agent scaffold, the retry policy, and the episode
composition of \Cref{tab:evaluation-protocol}. Every clean episode is paired
with one perturbed episode per family, and both sides of a pair are scored by
the benchmark's native grader. Beyond the interaction cap and the episode
timeout listed there, the two benchmarks differ in how the grader reports an
outcome and in how their task sets were assembled; we describe both below.

\begin{table*}[t]
    \centering
    \caption{\textbf{Evaluation protocol.} The upper block lists the controls
    shared by every run, and the lower block lists the two limits that differ
    between benchmarks.}
    \label{tab:evaluation-protocol}
    \footnotesize
    \setlength{\tabcolsep}{5pt}
    \renewcommand{\arraystretch}{1.2}
    \begin{tabularx}{\linewidth}{@{}>{\columncolor{appblue}\bfseries}p{0.26\linewidth}
        >{\raggedright\arraybackslash}X
        >{\raggedright\arraybackslash}X@{}}
        \toprule
        \textbf{Parameter} & \textbf{APEX-Agents} & \textbf{Toolathlon} \\
        \midrule
        \multicolumn{3}{@{}l}{\cellcolor{appsage}\textbf{Shared across
        benchmarks}} \\
        Agent scaffold & \multicolumn{2}{@{}p{0.7\linewidth}@{}}{Codex CLI,
        native function calling} \\
        System prompt & \multicolumn{2}{@{}p{0.7\linewidth}@{}}%
        {\Cref{app:prompts}} \\
        Retry policy & \multicolumn{2}{@{}p{0.7\linewidth}@{}}{3 attempts with
        exponential backoff from 30\,s, on rate-limit and timeout errors} \\
        Success oracle & \multicolumn{2}{@{}p{0.7\linewidth}@{}}{Native
        grader} \\
        Clean episodes & \multicolumn{2}{@{}p{0.7\linewidth}@{}}{25} \\
        Perturbations per episode & \multicolumn{2}{@{}p{0.7\linewidth}@{}}{12,
        one per family} \\
        \midrule
        \multicolumn{3}{@{}l}{\cellcolor{appamber}\textbf{Benchmark-specific}} \\
        Max. interaction turns & \textcolor{appmuted}{--} & 200 \\
        Episode timeout & 1800\,s & Agent timeout $+$ 600\,s \\
        \bottomrule
    \end{tabularx}
\end{table*}

\paragraph{Scoring conventions.}
The two graders return different kinds of quantities, so we keep each
benchmark on its native scale. APEX-Agents
returns a continuous reward in $[0,1]$ that awards partial credit when a
deliverable is partly correct, so a model's score is the mean reward over
episodes. Toolathlon returns a binary outcome per episode, so its mean reward
equals its success rate. Because partial credit and end-to-end success lie on
different scales, cross-benchmark comparisons use only relative degradation
(\Cref{sec:rq4-capability}), which normalizes by a model's own clean score.
Within a benchmark, absolute differences are reported directly. Unscored runs are excluded from aggregates.

\paragraph{Task sets.}
APEX-Agents contributes all 25 of its tasks. Toolathlon defines 79 tasks, of
which we report a subset of 25; the remaining tasks did not complete both the
clean episode and the full twelve-family perturbed sweep within our compute
budget. The subset is determined by run completion, so absolute Toolathlon
scores in this paper differ from published full-benchmark numbers. All
comparisons are paired, with the clean and perturbed sides of every contrast
drawn from the same 25 tasks.

\subsection{External Capability Proxy}
\label{app:capability-proxy}

For the exploratory analysis in \Cref{sec:rq4-capability}, we use the highest
published Artificial Analysis Intelligence Index v4.1.1 score for each model
family, selecting maximum effort when multiple effort levels are reported.
\Cref{tab:capability-proxy} lists the resulting inputs. We use family-level
proxies because the evaluation records do not identify an effort tier for
every model. Scores were retrieved on
September 29, 2026; the released metadata retain the per-model source URLs.

\begin{table}[!htb]
    \centering
    \caption{\textbf{External capability scores.} Scores are Artificial Analysis
    Intelligence Index v4.1.1 family-level proxies used in
    \Cref{fig:rq4-capability}. The last column gives the configuration from
    which each proxy is taken.}
    \label{tab:capability-proxy}
    \scriptsize
    \setlength{\tabcolsep}{3.5pt}
    \renewcommand{\arraystretch}{1.08}
    \begin{tabularx}{\columnwidth}{@{}>{\raggedright\arraybackslash}X
        >{\centering\arraybackslash}p{0.13\columnwidth}
        >{\raggedright\arraybackslash}p{0.38\columnwidth}@{}}
        \toprule
        \textbf{Model} & \textbf{AA Index} & \textbf{Proxy configuration} \\
        \midrule
        GPT-5.6-Sol       & 61 & Family max.; max effort \\
        Claude Opus 5     & 63 & Family max.; max effort \\
        DeepSeek-V4-Pro   & 53 & Reasoning; max effort \\
        DeepSeek-V4-Flash & 52 & Reasoning; max effort \\
        Kimi-K3           & 60 & Family max.; max effort \\
        Kimi-K2.6         & 45 & Reasoning \\
        Hunyuan-3-High    & 42 & Published Hy3 entry \\
        Qwen3.8-Max       & 58 & Published Qwen3.8-Max entry \\
        GLM-5.2           & 53 & Family max.; max effort \\
        GLM-5.1           & 41 & Reasoning \\
        \bottomrule
    \end{tabularx}
\end{table}

\subsection{Failure-Attribution Annotation Scheme}
\label{app:attribution}

This subsection defines the labels used in \Cref{sec:awareness}, and complete
sample-level annotations are included in the released artifacts. We examine
all 30 matched \deepseekicon DeepSeek-V4-Flash pairs in which the clean
episode succeeds and the perturbed episode fails. These comprise 20
Toolathlon and 10 APEX-Agents pairs spanning 13 tasks.

Each complete reasoning and action trace receives two complementary labels.
$A$ records the highest explicit awareness grade reached by the agent, and
$E$ records its immediate behavior at the first observable error. The
awareness grades are nested, and only the highest grade reached is assigned.
Candidate expressions of doubt are retrieved with an epistemic-marker lexicon,
and each candidate is then judged in context. Behavior labels describe what
is observable in the trace and distinguish unsupported content, omissions, and action-level defects,
following prior work on factuality and faithfulness
\citep{ji2023hallucination,maynez2020faithfulness,kryscinski2020factcc,
huang2025hallucination,dziri2023faith,kadavath2022know}.

\begin{table*}[t]
    \centering
    \caption{\textbf{Failure-attribution labels.} Awareness grades (A1--A4) capture
    only signals verbalized in the trace, and behavior labels (E1--E6) are
    assigned at the first observable error.}
    \label{tab:error-behavior-definitions}
    \footnotesize
    \setlength{\tabcolsep}{4pt}
    \renewcommand{\arraystretch}{1.15}
    \begin{tabularx}{\linewidth}{@{}>{\centering\arraybackslash}p{0.10\linewidth}
        >{\centering\arraybackslash}p{0.06\linewidth}
        >{\raggedright\arraybackslash}p{0.22\linewidth}
        >{\raggedright\arraybackslash}X@{}}
        \toprule
        \textbf{Axis} & \textbf{Code} & \textbf{Label} &
        \textbf{Operational definition} \\
        \midrule
        \multirow{4}{*}{Awareness}
        & A1 & No explicit signal &
        The trace contains no explicit indication that a value, conclusion,
        or answer may be problematic. \\
        & A2 & Tentative signal &
        The trace contains a tentative or one-off indication that a value,
        conclusion, or answer may be problematic. \\
        & A3 & Sustained problem statement &
        The agent explicitly and persistently identifies a specific value,
        conclusion, or answer as problematic. \\
        & A4 & Input-channel attribution &
        The agent attributes the problem to the integrity of an instruction,
        context source, or tool channel. \\
        \midrule
        \multirow{6}{*}{Behavior}
        & E1 & Invented content &
        The agent supplies an identifier, value, or other content unsupported
        by the available evidence. \\
        & E2 & Requirement omitted &
        Required content or an output constraint is absent from the
        deliverable. \\
        & E3 & Corrupted value propagated &
        The agent uses a corrupted source or intermediate value and carries it
        into downstream reasoning or the final output. \\
        & E4 & Task abandoned &
        The agent stops after concluding that it cannot proceed, without
        clarification or recovery. \\
        & E5 & Code defect &
        Relevant reasoning is followed by faulty data-processing code or an
        inconsistent operation or artifact. \\
        & E6 & Judgment reversed &
        The agent reaches a professional judgment opposite to the conclusion
        supported by the available evidence. \\
        \bottomrule
    \end{tabularx}
\end{table*}

\section{Additional Analyses}
\label{app:additional-results}

\subsection{Tool-Perturbation Edit Magnitude}
\label{app:tool-ablation}

\takeaway{For Hunyuan-3-High, larger tool edits do not cause larger losses,
and the smallest edit, description replacement, is the most damaging.}

We analyze 67 tool-targeted perturbations for \hunyuanmodelicon
Hunyuan-3-High, grouped by patch operation from field-level edits to
whole-tool changes. Each perturbation is paired with the clean outcome of its
base task. Success drop is the mean paired clean-minus-perturbed difference
within each group, with positive values indicating degradation. Of the 25
clean base tasks, 16 are successful.

\begin{table}[H]
    \centering
    \caption{\textbf{Effects of tool-contract edits by patch operation.} Results
    are for Hunyuan-3-High. Drop is the paired success drop in percentage
    points against each base task's clean outcome, and $\Delta$ cost compares
    perturbed and clean mean token cost among successful trajectories. Color
    intensity encodes drop magnitude.}
    \label{tab:tool-edit-levels}
    \scriptsize
    \setlength{\tabcolsep}{3pt}
    \renewcommand{\arraystretch}{1.22}
    \begin{tabularx}{\columnwidth}{@{}l
        >{\raggedright\arraybackslash}Xccr@{}}
        \toprule
        \rowcolor{appblue!75}
        \textbf{Level} & \textbf{Patch operation} & $\bm{n}$ &
        \textbf{Drop} & $\bm{\Delta}$\,\textbf{cost} \\
        \midrule
        \levelbadge{levelone}{L1} &
        \texttt{replace\_description} & 19 &
        \cellcolor{dropred!24}\textcolor{dropred}{\bfseries 21.1} &
        \textcolor{ratioincrease}{\bfseries +97,616} \\
        \levelbadge{leveltwo}{L2} &
        \texttt{replace\_enum} & 6 &
        \cellcolor{dropred!19}\textcolor{dropred}{\bfseries 16.7} &
        \textcolor{deltagreen}{\bfseries $-2,452$} \\
        \levelbadge{levelthree}{L3} &
        \texttt{add\_fake\_tool} & 23 &
        \cellcolor{dropred!7}\textcolor{dropred}{\bfseries 4.3} &
        \textcolor{deltagreen}{\bfseries $-19,691$} \\
        \levelbadge{levelfour}{L4} &
        \texttt{remove\_tool} & 19 &
        \cellcolor{dropred!8}\textcolor{dropred}{\bfseries 5.3} &
        \textcolor{ratioincrease}{\bfseries +11,652} \\
        \bottomrule
    \end{tabularx}
\end{table}

\Cref{tab:tool-edit-levels} shows that the losses do not grow with edit
magnitude. Description replacement causes the largest success drop and the
largest cost increase among successful trajectories, whereas adding a fake
tool or removing a tool costs only about five points. A false description
leaves the interface callable and raises no execution error, while structural
inconsistencies are conspicuous enough to be detected or bypassed. Group sizes
and task composition differ across levels, so these comparisons are
descriptive.

\begin{table}[H]
    \centering
    \caption{\textbf{Representative tool-contract edits.} Each row shows one patch
    per edit level, compressed to the field that changes, and gray text gives
    the source perturbation.}
    \label{tab:tool-edit-examples}
    \scriptsize
    \setlength{\tabcolsep}{4pt}
    \renewcommand{\arraystretch}{1.18}
    \begin{tabularx}{\columnwidth}{@{}l>{\raggedright\arraybackslash}X@{}}
        \toprule
        \rowcolor{appblue!75}
        \textbf{Level} & \textbf{Representative edit} \\
        \midrule
        \levelbadge{levelone}{L1} &
        Rewrite the description of \texttt{read\_text\_file} to claim that only
        listed file extensions are readable.\newline
        \textcolor{appmuted}{\texttt{apex-0097f20d/T\_mistake}} \\
        \levelbadge{leveltwo}{L2} &
        Remove \texttt{list} from the \texttt{calendar} action enum.\newline
        \textcolor{appmuted}{\texttt{apex-075c6f8f/T\_missing}} \\
        \levelbadge{levelthree}{L3} &
        Add a nonexistent \texttt{zip\_folder} tool with a plausible
        compression interface.\newline
        \textcolor{appmuted}{\texttt{apex-0097f20d/T\_redundant}} \\
        \levelbadge{levelfour}{L4} &
        Remove \texttt{read\_text\_file} from the advertised tool set.\newline
        \textcolor{appmuted}{\texttt{apex-0097f20d/T\_missing}} \\
        \bottomrule
    \end{tabularx}
\end{table}

\subsection{Cross-Family Sensitivity Correlations}
\label{sec:rq3-correlation}

\begin{figure*}[t]
    \centering
    \includegraphics[width=0.86\linewidth]{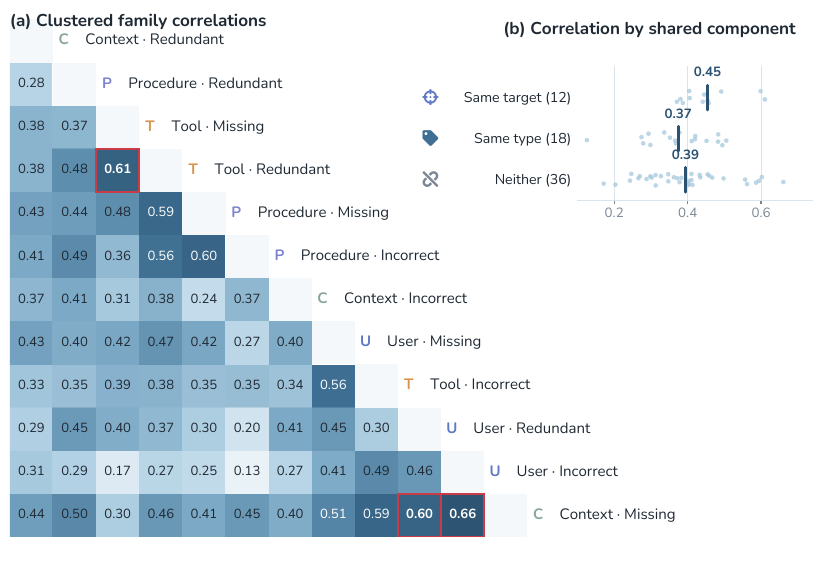}
    \caption{\textbf{Correlations between perturbation families.} \textbf{(a)}
    Partial Spearman correlations of relative degradation across $n=75$
    model--task pairs, controlling for clean performance and model identity.
    \textbf{(b)} The 66 pairwise correlations grouped by the component the two
    families share, with vertical bars marking group means.}
    \label{fig:rq3-correlation}
\end{figure*}

\begin{table*}[t]
    \centering
    \caption{\textbf{Rollout length under perturbation.} Entries are median
    perturbed-to-clean ratios with first and third quartiles in brackets, so
    values below one indicate shorter perturbed rollouts. Blue and orange
    medians mark decreases and increases, shaded entries denote nominal
    significance ($p<.05$), and $n$ is the number of matched trajectory
    pairs.}
    \label{tab:rollout-length}
    \scriptsize
    \setlength{\tabcolsep}{2.7pt}
    \renewcommand{\arraystretch}{1.14}
    \begin{tabularx}{\linewidth}{@{}>{\raggedright\arraybackslash}Xccccc@{}}
        \toprule
        \rowcolor{appblue!75}
        \textbf{Group} & \textbf{Tool calls} & \textbf{Turns} &
        \textbf{Input tokens} & \textbf{Output tokens} & $\bm{n}$ \\
        \midrule
        \rowcolor{appblue!35}
        \multicolumn{6}{@{}l}{\textit{By perturbation type}} \\
        \apptypemissing
        & \ratiodown{0.89}{0.66, 1.37} & \ratiodown{0.87}{0.62, 1.30}
        & \rationeutral{1.00}{0.75, 1.24} & \rationeutral{1.02}{0.72, 1.44} & 96 \\
        \apptypeincorrect
        & \ratiodown{0.90}{0.67, 1.22} & \rationeutral{1.00}{0.63, 1.18}
        & \ratiodown{0.92}{0.64, 1.17} & \rationeutral{0.98}{0.68, 1.47} & 93 \\
        \apptyperedundant
        & \ratiodown{0.93}{0.56, 1.43} & \ratiodown{0.91}{0.60, 1.42}
        & \ratiodown{0.91}{0.57, 1.27} & \rationeutral{0.99}{0.72, 1.39} & 90 \\
        \midrule
        \rowcolor{appblue!35}
        \multicolumn{6}{@{}l}{\textit{By perturbation target}} \\
        \apptargetuser\enspace User utterance
        & \ratiodown{0.93}{0.67, 1.10} & \ratiosigdown{0.82}{0.63, 1.18}
        & \ratiodown{0.92}{0.62, 1.11} & \ratiodown{0.91}{0.71, 1.30} & 69 \\
        \apptargetprocedure\enspace Procedure
        & \ratiodown{0.93}{0.61, 1.43} & \rationeutral{0.95}{0.59, 1.43}
        & \rationeutral{0.95}{0.71, 1.23} & \ratioup{1.07}{0.71, 1.38} & 69 \\
        \apptargetcontext\enspace Context
        & \rationeutral{0.99}{0.66, 1.43} & \rationeutral{0.97}{0.65, 1.44}
        & \rationeutral{0.95}{0.67, 1.38} & \ratiosigup{1.11}{0.77, 1.66} & 72 \\
        \apptargettool\enspace Tool contract
        & \ratiodown{0.83}{0.66, 1.33} & \ratiodown{0.88}{0.62, 1.25}
        & \ratiodown{0.94}{0.58, 1.18} & \ratiodown{0.93}{0.69, 1.35} & 69 \\
        \bottomrule
    \end{tabularx}
\end{table*}

\takeaway{Sensitivity is organized more strongly by where information is
perturbed than by how it is perturbed.}

To characterize whether different perturbations induce vulnerability on the
same task instances, we compute partial Spearman correlations between the
performance-degradation profiles of all 12 families, controlling for clean
performance and model identity. The resulting correlations in
\Cref{fig:rq3-correlation} are moderate, indicating that different families
affect partially overlapping model--task pairs. Perturbations applied to the
same target are more strongly correlated than those that share only a
perturbation type, so model sensitivity is organized primarily by where
information is disturbed.

\paragraph{Correlation estimator.}
For model $m$, task $i$, and perturbation family $f$, we normalize the signed
performance change by the model's mean clean score,
\[
    d^{(f)}_{mi}
    =
    \frac{\widetilde{c}_{mi}-\widetilde{p}^{(f)}_{mi}}
         {\frac{1}{N}\sum_j \widetilde{c}_{mj}}.
\]
We rank these changes within each family and regress out clean-performance
rank and model fixed effects:
\[
    \begin{aligned}
    R(d^{(f)}_{mi})
    &= \alpha_f + \beta_f R(\widetilde{c}_{mi}) \\
    &\quad + \sum_{m'}\gamma_{fm'}\mathbb{I}[m=m']
    + \varepsilon^{(f)}_{mi}.
    \end{aligned}
\]
The reported partial Spearman correlation is
$\rho_{fg}=\operatorname{Corr}(\varepsilon^{(f)},\varepsilon^{(g)})$.
Families are ordered for display by average-linkage clustering on
$1-\rho_{fg}$.

\paragraph{Correlation intervals.}
We quantify the uncertainty of the partial Spearman correlations in
\Cref{fig:rq3-correlation} with a cluster bootstrap over tasks. Each of the
2000 replicates resamples the 25 tasks with replacement, keeps all models of a
resampled task together because their scores share one episode, and repeats
the entire estimation, including the rank regression on clean score and model
identity. \Cref{tab:rq3-intervals} reports percentile intervals for the three
strongest pairs, the weakest pair, and the grouping by shared component.

Individual-pair intervals are broad, with a median width of 0.51, yet 58 of
the 66 exclude zero, indicating substantial shared sensitivity. The grouped
comparison is also stable, as both contrasts between pairs that share a target
and the other two groups exclude zero.

\begin{table}[H]
    \centering
    \caption{\textbf{Bootstrap intervals for family correlations.} Entries are
    cluster-bootstrap percentile intervals for the partial Spearman
    correlations of \Cref{fig:rq3-correlation}. The upper block gives the
    three strongest and the weakest family pair, the middle block groups all
    66 pairs by what the two families share, and the lower block reports the
    contrasts between these groups.}
    \label{tab:rq3-intervals}
    \scriptsize
    \setlength{\tabcolsep}{3pt}
    \renewcommand{\arraystretch}{1.15}
    \begin{tabular}{@{}>{\raggedright\arraybackslash}p{0.58\columnwidth}
        >{\centering\arraybackslash}p{0.10\columnwidth}
        >{\centering\arraybackslash}p{0.24\columnwidth}@{}}
        \toprule
        \textbf{Quantity} & $\bm{\rho}$ & \textbf{95\% CI} \\
        \midrule
        Context-Missing \& User-Incorrect & 0.66 & $[0.50,\,0.81]$ \\
        Tool-Redundant \& Tool-Missing & 0.61 & $[0.41,\,0.79]$ \\
        Context-Missing \& User-Redundant & 0.60 & $[0.44,\,0.77]$ \\
        Procedure-Incorrect \& User-Incorrect & 0.13 & $[-0.18,\,0.46]$ \\
        \midrule
        Same target (\textit{n}\,=\,12) & 0.45 & $[0.35,\,0.56]$ \\
        Same type (\textit{n}\,=\,18) & 0.37 & $[0.26,\,0.49]$ \\
        Neither (\textit{n}\,=\,36) & 0.39 & $[0.27,\,0.51]$ \\
        \midrule
        Same target $-$ same type & +0.08 & $[0.00,\,0.15]$ \\
        Same target $-$ neither & +0.06 & $[0.00,\,0.13]$ \\

        \bottomrule
    \end{tabular}
\end{table}

\subsection{Effects on Rollout Length and Stability}
\label{app:rollout-length}

\takeaway{Perturbations leave the typical rollout length nearly unchanged but
increase tail risk, occasionally inducing loop-like trajectories with
exceptionally high token consumption.}

We examine whether perturbations systematically change the number of tool
calls, interaction turns, input tokens, and output tokens in a rollout. For each matched pair, we divide the perturbed value by its
clean counterpart, so a ratio of one indicates no change. Because a small
number of loop-like trajectories produce heavy-tailed ratios, we use the
median and interquartile range as the primary summary and treat means and
standard deviations as diagnostics of tail instability.

\Cref{tab:rollout-length} shows little systematic change in typical rollout
length, with every median ratio between 0.82 and 1.11. Only two of the 28
comparisons reach nominal significance. User-utterance perturbations slightly
shorten interactions, and context perturbations slightly lengthen outputs.
With this many comparisons, neither effect is strong evidence.

The tails tell a different story. For \textsc{Incorrect} and context
perturbations, the mean input-token ratio approaches two and its standard
deviation is close to six, even though both medians stay below one. A few
loop-like trajectories account for this spread, while typical rollouts stay
close to their clean length.

\section{Prior Robustness Evaluations}
\label{app:prior-evaluation-coverage}

\Cref{tab:related-coverage} maps representative robustness evaluations onto
our targets and perturbation types as a qualitative scope comparison.

\newcommand{\rwyes}{\textcolor{deltagreen}{\faCheck}}
\newcommand{\rwpartial}{\textcolor{orange}{\(\triangle\)}}
\newcommand{\rwno}{\textcolor{gray}{--}}
\begin{table}[H]
    \centering
    \caption{\textbf{Coverage of prior robustness evaluations.} A check marks
    direct coverage and a triangle a partial or analogous match, expressed in
    our terminology. $T$ denotes the static tool contract rather than runtime
    tool outputs, so interventions on observations or environment state may
    have type coverage without a target check.}
    \label{tab:related-coverage}
    \scriptsize
    \setlength{\tabcolsep}{2.4pt}
    \renewcommand{\arraystretch}{1.13}
    \begin{tabularx}{\columnwidth}{@{}>{\raggedright\arraybackslash}X*{7}{c}@{}}
        \toprule
        \multirow{2}{*}{\textbf{Evaluation}} &
        \multicolumn{4}{c}{\textbf{Perturbation target}} &
        \multicolumn{3}{c}{\textbf{Type}} \\
        \cmidrule(lr){2-5}\cmidrule(l){6-8}
        & $U$ & $P$ & $C$ & $T$
        & \textbf{Miss.} & \textbf{Red.} & \textbf{Inc.} \\
        \midrule
        \rowcolor{groupgray}
        \multicolumn{8}{@{}l}{\textbf{Foundation-model robustness}} \\
        PromptRobust
            & \rwpartial & \rwno & \rwno & \rwno
            & \rwno & \rwyes & \rwpartial \\
        SCORE
            & \rwpartial & \rwno & \rwno & \rwno
            & \rwno & \rwno & \rwno \\
        Math-RoB
            & \rwpartial & \rwno & \rwno & \rwno
            & \rwyes & \rwyes & \rwno \\
        \addlinespace[2pt]
        \rowcolor{groupgray}
        \multicolumn{8}{@{}l}{\textbf{Agentic robustness}} \\
        AgentDojo
            & \rwno & \rwno & \rwno & \rwno
            & \rwno & \rwno & \rwno \\
        ReliabilityBench
            & \rwyes & \rwno & \rwno & \rwpartial
            & \rwyes & \rwyes & \rwpartial \\
        AgentNoiseBench
            & \rwyes & \rwno & \rwno & \rwno
            & \rwyes & \rwyes & \rwyes \\
        AgentHijack
            & \rwno & \rwno & \rwno & \rwno
            & \rwpartial & \rwyes & \rwno \\
        ToolMaze
            & \rwno & \rwno & \rwno & \rwno
            & \rwyes & \rwno & \rwyes \\
        \rowcolor{abstractbg}
        \textbf{Ours}
            & \rwyes & \rwyes & \rwyes & \rwyes
            & \rwyes & \rwyes & \rwyes \\
        \bottomrule
    \end{tabularx}
\end{table}

\section{Perturbation Examples}
\label{app:examples}

\Cref{tab:example-gallery} instantiates all 12 families in a single
repository-coding task, so that the families differ only in the target that is
edited and in the type of edit applied to it.

\begin{table}[H]
    \centering
    \caption{\textbf{Examples of the 12 perturbation families.} All examples are
    set in one repository-coding task, and each group changes only the listed
    target.}
    \label{tab:example-gallery}
    \scriptsize
    \setlength{\tabcolsep}{4pt}
    \renewcommand{\arraystretch}{1.12}
    \begin{tabularx}{\columnwidth}{@{}>{\centering\arraybackslash}p{0.11\columnwidth}
        >{\raggedright\arraybackslash}X@{}}
        \toprule
        \textbf{Target} & \textbf{Schematic edits} \\
        \midrule
        \cellcolor{appblue}\textbf{$U$} &
        \textbf{Missing:} remove the requested cache TTL.\newline
        \textbf{Redundant:} append an earlier deployment request after a topic
        switch.\newline
        \textbf{Incorrect:} change the target branch from \texttt{main} to
        \texttt{release}. \\[2pt]
        \cellcolor{appsage}\textbf{$P$} &
        \textbf{Missing:} omit the required approval condition.\newline
        \textbf{Redundant:} retain an obsolete review workflow beside the
        current policy.\newline
        \textbf{Incorrect:} replace ``test before merge'' with ``merge before
        test.'' \\[2pt]
        \cellcolor{appamber}\textbf{$C$} &
        \textbf{Missing:} drop the earlier traceback needed for diagnosis.\newline
        \textbf{Redundant:} carry over discussion from an unrelated
        repository.\newline
        \textbf{Incorrect:} attribute a configuration value to the wrong
        module. \\[2pt]
        \cellcolor{appblue}\textbf{$T$} &
        \textbf{Missing:} hide a required tool argument.\newline
        \textbf{Redundant:} add duplicate aliases for the same repository
        operation.\newline
        \textbf{Incorrect:} expose \texttt{search-code} while the server registers
        \texttt{search\_code}. \\
        \bottomrule
    \end{tabularx}
\end{table}

\begingroup
\def\renderappendixcase{1}

\endgroup

\section{Prompt Templates}
\label{app:prompts}

The generation code implements one shared prompt template with
target-specific slot ontologies, edit strategies, and output contracts. Each
variant is produced by one LLM call over the original target content. The
harness parses the returned JSON and writes the replacement into a copied
variant directory; the generator never edits task files directly. Grader
files under \texttt{tests/} and the \texttt{[verifier]} configuration remain
pristine. Every template below follows the same five-part schema of role,
input, task, constraints, and output. Each role is written as the persona
whose text is being produced, so the model writes in the voice of the
original author.

\subsection{Prompt Assembly and Label Mapping}

The implementation uses \texttt{mistake} as the internal name for
\textsc{Incorrect}. The four code-level positions map one-to-one to our paper
notation as follows. All four targets apply to MCP tasks; coding tasks expose
only the user-instruction surface in the current implementation.

\begin{table*}[t]
    \centering
    \caption{\textbf{Target-to-generator mapping.} Each row lists the generator
    input, source surface, returned replacement, and applicable domain for
    one perturbation target.}
    \label{tab:prompt-target-mapping}
    \footnotesize
    \setlength{\tabcolsep}{3.5pt}
    \renewcommand{\arraystretch}{1.2}
    \begin{tabularx}{\linewidth}{@{}c
        >{\ttfamily\raggedright\arraybackslash}p{0.16\linewidth}
        >{\raggedright\arraybackslash}p{0.25\linewidth}
        >{\raggedright\arraybackslash}Xl@{}}
        \toprule
        \textbf{Target} & \textbf{attack\_face} & \textbf{Source surface} &
        \textbf{Returned replacement} & \textbf{Domain} \\
        \midrule
        \cellcolor{appblue}$U$ & instruction & \texttt{instruction.md} &
        Full rewritten instruction & Coding, MCP \\
        \cellcolor{appsage}$P$ & plan & Rough plan / skill note &
        Markdown appended to instruction & MCP \\
        \cellcolor{appamber}$C$ & context & Synthetic prior conversation &
        Serialized role--content turns & MCP \\
        \cellcolor{appblue}$T$ & tool\_schema & MCP \texttt{tools/list} &
        Serialized schema patch operations & MCP \\
        \bottomrule
    \end{tabularx}
\end{table*}

\begin{table*}[t]
    \centering
    \caption{\textbf{Corruption semantics.} Each row gives the instruction injected
    into the shared generation prompt for one perturbation type. The
    implementation label \texttt{mistake} corresponds to \textsc{Incorrect}.}
    \label{tab:prompt-corruption-semantics}
    \footnotesize
    \setlength{\tabcolsep}{5pt}
    \renewcommand{\arraystretch}{1.2}
    \begin{tabularx}{\linewidth}{@{}>{\bfseries\color{appnavy}}p{0.16\linewidth}X@{}}
        \toprule
        Missing & Quietly remove information that is genuinely needed, rather
        than decorative, so that an agent that follows the surface blindly
        cannot complete the task. \\
        \rowcolor{appblue!45}
        Incorrect & Introduce a realistic human-like orthographic,
        naming-convention, or configuration error that remains recoverable
        without changing the task's true answer. \\
        Redundant & Keep the truth intact while adding plausible but irrelevant
        material, ghost entries, duplicates, or verbose detail that dilutes
        attention. \\
        \bottomrule
    \end{tabularx}
\end{table*}

\subsection{Target and Type Selection Prompt}
\label{app:selection-prompt}

We use DeepSeek-V4-Pro to select a compatible perturbation target and type for
each clean benchmark instance. This step selects the intervention but does not
modify the instance.

\begin{promptcard}{Target and type selection template}
    \promptlabel{Role}
    You are a benchmark engineer preparing a robustness study. You read a clean
    agentic task and decide where it could be disturbed and in what way. You
    are a planner, not an editor: you never touch the task itself.

    \promptlabel{Input}
    {\raggedright
    \slot{[BENCHMARK DESCRIPTION]}\hspace{3pt}
    \slot{[CLEAN INSTANCE]}\hspace{3pt}
    \slot{[BENCHMARK METADATA]}\par}

    \promptlabel{Task}
    Choose exactly one target and exactly one perturbation type. A target is
    eligible only when the instance actually carries that information.

    \vspace{2pt}
    \slotline{U}{the request the user states}
    \slotline{P}{procedures, policies, skill notes}
    \slotline{C}{dialogue history, summaries, memory}
    \slotline{T}{tool names, argument schemas, observation fields}
    \vspace{3pt}
    \slotline{missing}{remove information the task needs}
    \slotline{redundant}{add plausible but irrelevant information}
    \slotline{incorrect}{replace valid information with false, inconsistent, or
    interface-incompatible content}

    \promptlabel{Constraints}
    \begin{promptrules}
        \item The pair must fit the native structure of the instance.
        \item The task goal stays the same and at least one path to completion
        remains.
        \item Do not modify or rewrite any part of the clean instance.
    \end{promptrules}

    \promptlabel{Output}
    \begin{promptjson}
        \{"target": "U|P|C|T",\\
        \hspace*{1em}"type": "missing|redundant|incorrect",\\
        \hspace*{1em}"reasoning": "1-3 sentences"\}
    \end{promptjson}

    \promptlabel{Final instruction}
    Return the JSON object and nothing else.
\end{promptcard}

\subsection{Shared Generation Prompt}
\label{app:generation-prompt}

\begin{promptcard}{Main perturbation-generation template}
    \promptlabel{Role}
    You are a task author for a robustness benchmark. You rewrite exactly one
    component of an agent task so that a careless agent follows the rewritten
    version while a careful agent notices that something is off. You write in
    the voice of whoever authored the original component, never in the voice of
    an evaluator.

    \promptlabel{Input}
    {\raggedright
    \slot{[POSITION LABEL]}\hspace{3pt}
    \slot{[SLOT ONTOLOGY]}\hspace{3pt}
    \slot{[CORRUPTION TYPE AND MEANING]}\hspace{3pt}
    \slot{[ORIGINAL CONTENT]}\hspace{3pt}
    \slot{[CONCRETE TRICKS]}\hspace{3pt}
    \slot{[OUTPUT CONTRACT]}\par}

    \promptlabel{Task}
    Rewrite the content between
    \texttt{<original\_content>} and \texttt{</original\_content>} using at
    least one strategy from the injected trick menu. That content is ground
    truth: nothing outside it may be invented, and nothing inside it may change
    except through the chosen strategy.

    \promptlabel{Constraints}
    \begin{promptrules}
        \item \textbf{Preserve the answer.} Numbers, dates, amounts, verdicts,
        entity identities, and output requirements such as ``top 3 regions''
        keep their meaning. They may carry orthographic noise but never a
        different value.
        \item \textbf{Keep identifiers recognizable.} Case and naming
        convention may change; the referent may not. Environment values such
        as URLs, ports, timeouts, and image hashes may break only when the
        correct answer survives.
        \item \textbf{Stay plausible.} Match the benchmark's native style;
        placeholders such as \texttt{foo\_bar\_baz} are forbidden.
        \item \textbf{Leave no trace.} Never use ``fake,'' ``corrupted,''
        ``attack,'' ``poison,'' ``benchmark,'' ``perturb,'' ``injected,'' or
        ``trap,'' and never hedge with ``I think,'' ``maybe,''
        ``double-check,'' or ``correct me if.''
        \item \textbf{Keep it valid.} TOML, Dockerfile, JSON, and Markdown
        syntax survive except for the deliberately changed bytes.
        \item For \textsc{Redundant}, every true fact stays, so the correct
        answer remains reachable.
    \end{promptrules}

    \promptlabel{Output}
    \begin{promptjson}
        \{"perturbed": "<full replacement, per the output contract>",\\
        \hspace*{1em}"target\_slots": ["<slot>", ...],\\
        \hspace*{1em}"poisoned\_terms": ["<misleading token>", ...],\\
        \hspace*{1em}"reasoning": "1-3 sentences"\}
    \end{promptjson}

    \promptlabel{Final instruction}
    Return the JSON object and nothing else.
\end{promptcard}

\subsection{Target-Specific Prompt Blocks}

\begin{appendixcard}{$U$ · User instruction (\texttt{instruction.md})}
    \targetbadge{$U$}\hspace{0.35em}
    \textcolor{appmuted}{\scriptsize Coding and MCP tasks}

    \promptlabel{Role}
    You are the user who wrote this request. You restate it in your own words,
    imperfectly, the way a hurried person would.

    \promptlabel{Slot ontology}
    Task facts such as entities, roles, statutes, rates, courts, dates, and
    amounts; output-format and length constraints; workspace paths and
    filenames; and, for MCP tasks, server names, tool categories, and action
    hints.

    \tcbline
    \typedesc{Missing}{Drop the answer-format contract, one essential fact, a
        required working directory, or an MCP server or tool-category hint,
        keeping the instruction grammatical.}
    \typedesc{Incorrect}{Add a recognizable typo or homophone, change only the
        case or naming convention of a tool, insert a natural self-correction,
        or add a short run of decodable phonetic noise.}
    \typedesc{Redundant}{Prepend at least 300 words of plausible backstory,
        repeat the core requirement two or three times, or digress about
        unrelated workflows and tool categories.}
\end{appendixcard}

\begin{appendixcard}{$P$ · Rough plan / skill note (\texttt{rough\_plan.md})}
    \targetbadge{$P$}\hspace{0.35em}
    \textcolor{appmuted}{\scriptsize MCP tasks}

    \promptlabel{Role}
    You are the engineer who drafted this plan. You revise your own notes, so
    the result still reads like working documentation rather than a rewrite.

    \promptlabel{Slot ontology}
    Ordered task steps; tool names, action verbs, argument names, and indexing
    conventions; and free-form skill notes or tutorial prose.

    \tcbline
    \typedesc{Missing}{Drop one essential step, the tool name a step depends
        on, or a required action verb or argument hint.}
    \typedesc{Incorrect}{Change a tool's case or naming convention, use a
        near-synonym that is not a valid action, state the wrong indexing
        convention, or swap an argument name for a near-equivalent.}
    \typedesc{Redundant}{Insert at least 300 words of unrelated tutorials,
        capability inventories, or tool catalogues, or bury duplicated correct
        steps in verbose digressions.}
\end{appendixcard}

\begin{appendixcard}{$C$ · Prior conversation context (\texttt{preamble.json})}
    \targetbadge{$C$}\hspace{0.35em}
    \textcolor{appmuted}{\scriptsize MCP tasks}

    \promptlabel{Role}
    You are transcribing the conversation that led to this request. You write
    both speakers, so the history must read as something that was said rather
    than something that was authored.

    \promptlabel{Construction}
    Because APEX-Agents provides no native dialogue history, the model first
    reverse-engineers a plausible sequence of user and assistant turns
    consistent with the live instruction, and only then perturbs it. The
    sequence is rendered above the instruction as ``Prior conversation with the
    user.''

    \tcbline
    \typedesc{Missing}{Remove one confirmed fact such as a path, entity, or
        negotiated output format, or delete the whole turn that carried the key
        requirement.}
    \typedesc{Incorrect}{Misspell or re-case an entity, path, or tool
        reference, or add a natural self-correction, without changing what is
        referred to.}
    \typedesc{Redundant}{Keep every task-relevant fact and add unrelated
        threads, small talk, tangents, or side quests that were already
        resolved.}
\end{appendixcard}

\begin{appendixcard}{$T$ · MCP tool schema (\texttt{tools/list})}
    \targetbadge{$T$}\hspace{0.35em}
    \textcolor{appmuted}{\scriptsize MCP tasks}

    \promptlabel{Role}
    You are the maintainer of this MCP server publishing a new revision of its
    schema. You emit machine-readable patch operations, not prose.

    \promptlabel{Slot ontology}
    Tool names and descriptions; meta-tool action enums; request-field names,
    types, descriptions, bounds, and indexing conventions; and the available
    tool set as a whole.

    \tcbline
    \typedesc{Missing}{Remove an essential tool, a required action from an
        enum, or a mandatory request field.}
    \typedesc{Incorrect}{Change only the spelling, case, or naming convention
        of a tool or field, misspell descriptive prose, or switch a documented
        indexing convention.}
    \typedesc{Redundant}{Keep every real tool and add three to six ghost or
        near-duplicate tools whose descriptions overlap with them.}
\end{appendixcard}

\subsection{Uniform JSON Output Contract}

\begin{promptcard}{Position-agnostic response envelope}
    Every generation call returns the same four-key envelope, whatever the
    target: \texttt{perturbed} carries the complete replacement,
    \texttt{target\_slots} lists the modified slots,
    \texttt{poisoned\_terms} records the injected misleading tokens and is
    empty for \textsc{Redundant}, and \texttt{reasoning} explains the edit. Only
    the payload of \texttt{perturbed} is target-specific.

    \tcbline
    \slotline{U}{Full rewritten \texttt{instruction.md}.}
    \slotline{P}{Full Markdown rough plan or skill note.}
    \slotline{C}{Serialized JSON array of \texttt{\{role, content\}} turns.}
    \slotline{T}{Serialized array of schema patch operations:
        \texttt{remove\_tool}, \texttt{replace\_description},
        \texttt{replace\_field\_description}, \texttt{replace\_enum},
        \texttt{replace\_type}, \texttt{remove\_field},
        \texttt{add\_fake\_tool}, or \texttt{prepend\_description}.}
\end{promptcard}

\subsection{Verification Prompt}
\label{app:verifier-prompt}

Every candidate is reviewed by a verifier LLM that does not participate in
generation. The four checks below are evaluated independently, and a candidate
is retained only when all four pass. The verifier decides the semantic checks;
schema validation and test execution are delegated to the benchmark's own
validators where they exist.

\begin{promptcard}{Candidate verification template}
    \promptlabel{Role}
    You are an independent auditor. You did not write the candidate and you
    will not fix it. You only decide whether the perturbed instance can serve
    as a matched evaluation pair for the clean one.

    \promptlabel{Input}
    {\raggedright
    \slot{[CLEAN INSTANCE]}\hspace{3pt}
    \slot{[PROPOSED EDIT]}\hspace{3pt}
    \slot{[DECLARED TARGET AND TYPE]}\hspace{3pt}
    \slot{[EVALUATION ANNOTATION]}\par}

    \promptlabel{Checks}
    \begin{promptrules}
        \item \textbf{Single target.} Only the declared target changed. The
        other three components are byte-identical, and grader files and
        verifier configuration are untouched. Any extra change is a
        \texttt{fail}, however harmless it looks.
        \item \textbf{Goal preservation.} The agent is asked for the same
        thing. Numbers, dates, amounts, entity identities, output
        requirements, and the expected answer or final environment state are
        unchanged. An edit that makes a different answer correct is a
        \texttt{fail}, not an \textsc{Incorrect} perturbation.
        \item \textbf{Solvability.} At least one interaction path still
        reaches the expected outcome. For \textsc{Missing}, say whether the
        omitted information is recoverable from the remaining components or
        through the clarification turn in the annotation. For
        \textsc{Redundant}, every fact needed for the solution is still
        present. For \textsc{Incorrect}, the error is detectable or repairable
        from within the episode.
        \item \textbf{Executability.} The instance still runs: container
        syntax is intact, declared tool calls resolve against the live server
        unless the tool contract is the target, and the annotation matches the
        benchmark's success oracle.
    \end{promptrules}

    \promptlabel{Constraints}
    \begin{promptrules}
        \item Judge only these four checks. Do not rate difficulty or
        interestingness.
        \item Do not repair the candidate. Anything that fails a check is
        rejected and regenerated.
        \item \texttt{decision} is \texttt{accept} only when all four checks
        pass; otherwise \texttt{reasoning} names the failed check.
    \end{promptrules}

    \promptlabel{Output}
    \begin{promptjson}
        \{"single\_target": "pass|fail",\\
        \hspace*{1em}"goal\_preserved": "pass|fail",\\
        \hspace*{1em}"solvable": "pass|fail",\\
        \hspace*{1em}"executable": "pass|fail",\\
        \hspace*{1em}"decision": "accept|reject",\\
        \hspace*{1em}"reasoning": "1-3 sentences"\}
    \end{promptjson}

    \promptlabel{Final instruction}
    Return the JSON object and nothing else.
\end{promptcard}

\end{document}